\documentclass[10pt,journal,compsoc]{IEEEtran}
\ifCLASSOPTIONcompsoc
  \usepackage[nocompress]{cite}
\else
  \usepackage{cite}
\fi
\usepackage[colorlinks,
linkcolor=green,       
anchorcolor=green,  
citecolor=green,        
]{hyperref}
\usepackage{ragged2e}
\usepackage[table]{xcolor} 
\usepackage{amsmath}
\usepackage{amsthm}
\usepackage{amssymb}
\usepackage{amsmath}
\usepackage{amssymb}
\usepackage{algorithm}
\usepackage{algorithmic}
\usepackage{bm} 
\usepackage{stfloats}
\usepackage{booktabs}   
\usepackage{multirow}   
\usepackage{pifont}     
\usepackage{amssymb}
\usepackage{booktabs} 
\usepackage{multirow} 
\usepackage{graphicx} 
\usepackage{amsmath}
\usepackage{xcolor}   
\usepackage{hyperref}
\usepackage{caption}
\usepackage{graphicx}
\usepackage[font=footnotesize,labelfont=sf,textfont=sf]{caption}
\usepackage{subcaption} 
\ifCLASSINFOpdf
\else
 
\fi      
\begin{document}
\title{Contrastive Parameter Disentanglement for Multi-modal Remote Sensing Image Generation}
\author{Yu~Zhang, Wenda~Zhao, Haojun~Tang, and Haipeng~Wang
\IEEEcompsocitemizethanks{
    
    \IEEEcompsocthanksitem \textit{Yu Zhang, Wenda Zhao and Haojun Tang are with the School of Information and Communication Engineering, Dalian University of Technology, Dalian 116024, China (e-mail: zyuiym@mail.dlut.edu.cn; zhaowenda@dlut.edu.cn; tanghaojun@mail.dlut.edu.cn).}\\
    
    \IEEEcompsocthanksitem \textit{H.Wang is with Unit 92728 of PLA,Shanghai, 200436, China. E-mail: whp5691@163.com.}
}%
} 
\IEEEtitleabstractindextext{%
\begin{abstract}
\justifying
 Remote sensing image generation is a pivotal technique for mitigating image scarcity in earth observation. However, existing generation methods are confined to single-modality synthesis, thereby failing to harness the complementary information inherent in multimodal images. In this paper, we propose a contrastive parameter disentanglement for multi-modal remote sensing image generation, which aims to generate semantically and structurally consistent multimodal remote sensing images (i.e., optical, infrared, and synthetic aperture radar) from a single textual prompt. The core challenge lies in mapping invariant semantics from a single text prompt while adapting to different modality attributes. We observe that LoRA adapters exhibit an inherent functional dichotomy, enabling disentanglement of semantic invariance and modality attributes at the parameter level. Thus, we propose a contrastive parameter disentanglement module that establishes the parameter-level disentanglement of shared semantics and distinct attributes within the orthogonal core subspace. After that, we build a disentangled optimization strategy that first constrains the parameter matrix A of the LoRA adapter to extract invariant semantics by a multi-modal contrastive objective, and then guides the multiple parameter matrices B to adapt different modality attributes under text prompts, thereby enabling the simultaneous generation of semantically consistent multimodal images. Furthermore, to ensure structural alignment across generated multimodal images, we devise a query-key structure transfer mechanism that enables joint modeling of multimodal sampling trajectories during inference by transferring structural correlation priors from the anchor modality to other modalities. Extensive experiments demonstrate that our proposed method outperforms state-of-the-art remote sensing image generation methods in terms of generation quality and semantic consistency, and yields superior performance in the downstream object classification task. Code will be released at https://github.com/DUT-ZYu/CPD-MMRS.
\end{abstract} 
}
\maketitle
\IEEEpeerreviewmaketitle
\ifCLASSOPTIONcompsoc
\IEEEraisesectionheading{\section{Introduction}\label{sec:introduction}}
\else
\section{Introduction}
\label{sec:introduction}
\fi
\IEEEPARstart{R}{emote} sensing image generation plays a vital role in promoting earth observation \cite{11298536,hong2024spectralgpt}, as it effectively alleviates the problem of image scarcity. Recently, generative models have achieved tremendous success and rapid development in text-to-image generation \cite{xing2025focus,guo2025conceptguard,10582893}, thereby further driving advances in remote sensing image generation. However, adapting these generative models to the remote sensing domain proves non-trivial, since overhead perspectives, large scales, and dense objects characterize remote sensing images. Consequently, recent studies have focused on customizing these models to the remote sensing domain. For instance, \cite{10988859,10768939} leverages the stable diffusion (SD) \cite{rombach2022high} model alongside global-scale datasets to achieve multi-resolution and unbounded synthesis. Meanwhile, other methods incorporate instance-level layouts \cite{11126950,11187367,yang2025mmo} or metadata \cite{sebaq2024rsdiff,khanna2024diffusionsat,sastry2024geosynth} to facilitate controllable generation. In essence, text-to-remote sensing image generation offers an intuitive interface for simulating complex earth observation scenarios via simple instructions, thereby expanding the application landscape of remote sensing.  \\
\begin{figure*}[!t]
\includegraphics[width=0.98\textwidth]{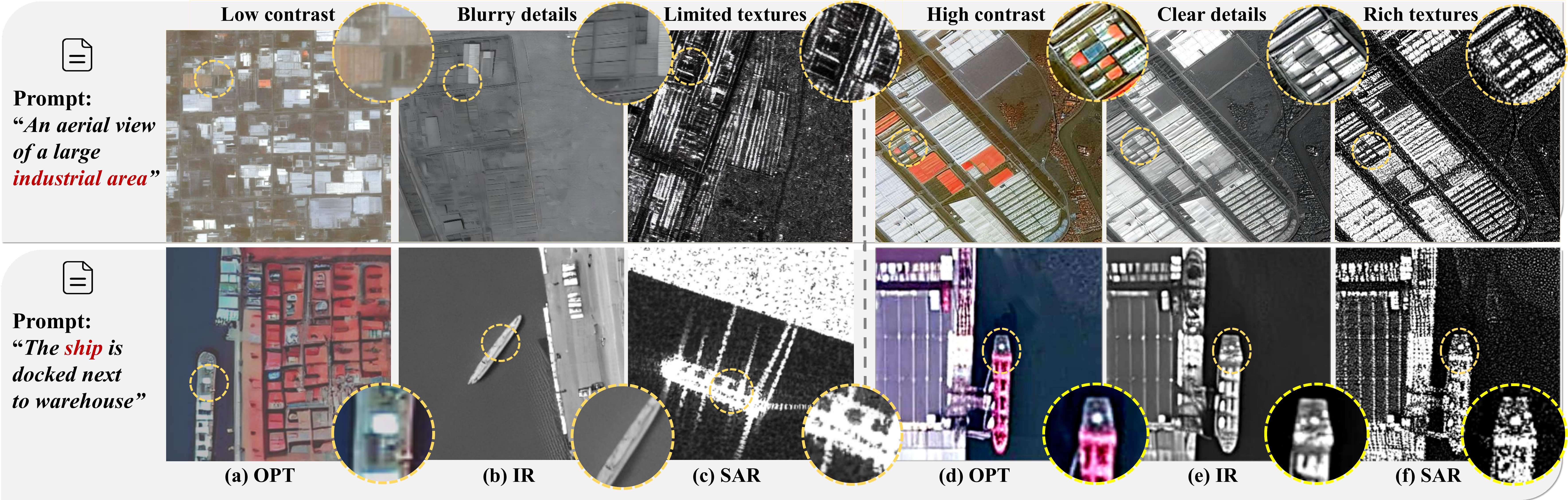}
		\centering 
		\caption{Comparison of the generated results. The multimodal images generated by Text2Earth \cite{10988859} (a-c) often face challenges such as low content contrast, blurred details, and limited textures. Conversely, the results generated by our proposed method (d-f) demonstrate high contrast, clear details, and rich textures from a single textual prompt, while maintains the cross-modal structural alignment.}
        \vspace{-10pt} 
		\label{fig1s}
	\end{figure*} 
\indent However, current research works are all limited to single-modal remote sensing image generation (e.g., optical (OPT) images), neglecting the complementary information provided by other modalities, such as infrared (IR) and synthetic aperture radar (SAR). As a result, such single-modal methods struggle to generate high-fidelity images across different modal types. As shown in Fig. \ref{fig1s}, the existing generation method Text2Earth\cite{10988859}, produces remote sensing images that lack fidelity in content contrast (e.g., OPT images), details (e.g., IR images), and textures (e.g., SAR images). In practical earth observation, these different modalities reflect diverse attributes and provide complementary perspectives. For instance, the OPT images provide fine structural and textural details via visible reflectance, yet is vulnerable to clouds and poor illumination. SAR images achieve reliable all-weather monitoring through active microwave backscattering, while IR images enable consistent day-and-night perception by capturing thermal radiation emitted from ground surfaces. Therefore, exploiting these multimodal images is crucial for boosting observation robustness under challenging conditions and enabling comprehensive synthesis of reliable remote sensing scenes.\\
\indent To this end, we propose a contrastive parameter disentanglement method for multi-modal remote sensing image
generation, aiming to bridge the research gap in multi-modal remote sensing synthesis. However, \emph{\textbf{how can a single textual prompt correctly map invariant semantics while also adapting to different modality attributes?}} A natural solution is to establish a text–semantic mapping via explicit semantic supervision, and then adapt modality attributes on top of the shared semantic representation. Unfortunately, strictly disentangling semantics from attributes is particularly challenging, as they are often intrinsically entangled. Existing methods in related fields mainly rely on reducing mutual information in latent spaces \cite{shan2024learning} or adopting independent encoders \cite{li2024learning}. Such approaches impose extrinsic constraints on latent representations while neglecting that semantic-attribute entanglement is inherently encoded in model parameters. Although recent low-rank adaptation (LoRA)-based methods \cite{hu2022lora}
(e.g., B-LoRA \cite{frenkel2024implicit} and ZipLoRA \cite{shah2024ziplora})
attempt to achieve disentanglement via layer-wise separation or orthogonal optimization between independent adapters, they fundamentally treat each LoRA module as an indivisible unit. Such a paradigm neglects that semantic-attribute entanglement is intrinsically embedded in the cross-layer parameters of LoRA, thus failing to fully exploit the parameter functional characteristics of LoRA.\\
\indent In this paper, we first analyze the parameter characteristic of the LoRA ($\Delta \mathbf{W} = \mathbf{B}\mathbf{A}$) by computing cross-modal cosine similarity across independently trained models. As shown in Fig. \ref{motivation}(a), we observe that different parameter matrices $\mathbf{A}$ exhibit strong cross-modal similarity, whereas different parameter matrices $\mathbf{B}$ show significant divergence. Such differences reveal that $\mathbf{A}$ naturally serves as a stable invariant anchor, while $\mathbf{B}$ specializes in encoding specific variations. This motivates us to anchor $\mathbf{A}$ to a stable semantic basis and let $\mathbf{B}$ freely adapt to diverse modal attributes. Moreover, implementing disentanglement in an unconstrained full parameter space is suboptimal, since it assumes that all parameter dimensions contribute equally to core representation. To investigate this, we conduct an dimensionality analysis across UNet layers of the SD model \cite{rombach2022high}. As visualized in Fig. \ref{motivation}(b), all layers exhibit a pervasive rank-deficient distribution, the core knowledge is encapsulated in a subspace, and the remaining dimensions correspond to redundancy. This indicates that optimization in the full parameter space impedes the accurate disentanglement. \\
\indent Based on the above analysis, we propose a contrastive parameter disentanglement (CPD) module, which achieves parameter-level disentanglement within the orthogonal core subspace by exploiting the inherent functional dichotomy of LoRA adapters so as to simultaneously extract invariant semantics and adapt modality attributes from a single text prompt. As illustrated in Fig. \ref{motivation}(c), CPD operates directly on the core parameter space instead of performing blind adaptation across the full parameter space, thereby mitigating redundant interference. Leveraging the structural independence of intrinsic orthogonal bases embedded in core parameter space, CPD explicitly delineates disentangled optimization directions for the parameter matrices $\mathbf{A}$ and $\mathbf{B}_{i}$ ($i\in (1,2,3)$). Guided by different basis directions, the parameter matrix $\mathbf{A}$ is contrast-constrained to disentangle cross-modal invariant semantics, while multiple independent parameter matrices $\mathbf{B}_{i}$ are optimized to adapt different modal attributes. Thus, our proposed CPD module can correctly map invariant semantics from a single textual prompt while also adapting to different modality attributes. \\
\begin{figure*}[t]
    \centering
    \begin{subfigure}[b]{0.27\linewidth}
        \centering
        \includegraphics[width=\linewidth]{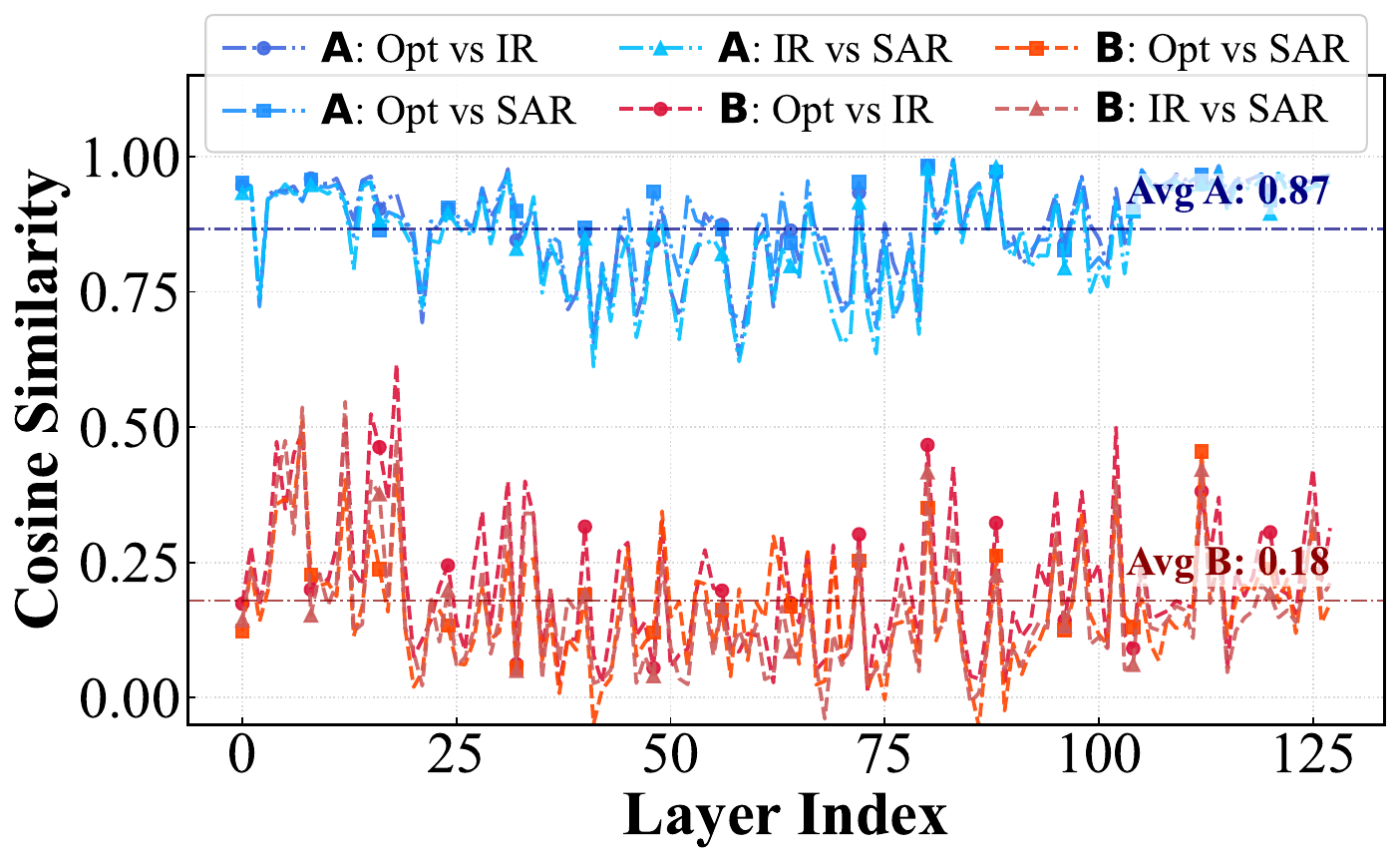}
        \caption{LoRA structure analysis}
        \label{fig:motivation_a}
    \end{subfigure}
    \hfill 
    \begin{subfigure}[b]{0.27\linewidth}
        \centering
        \includegraphics[width=\linewidth]{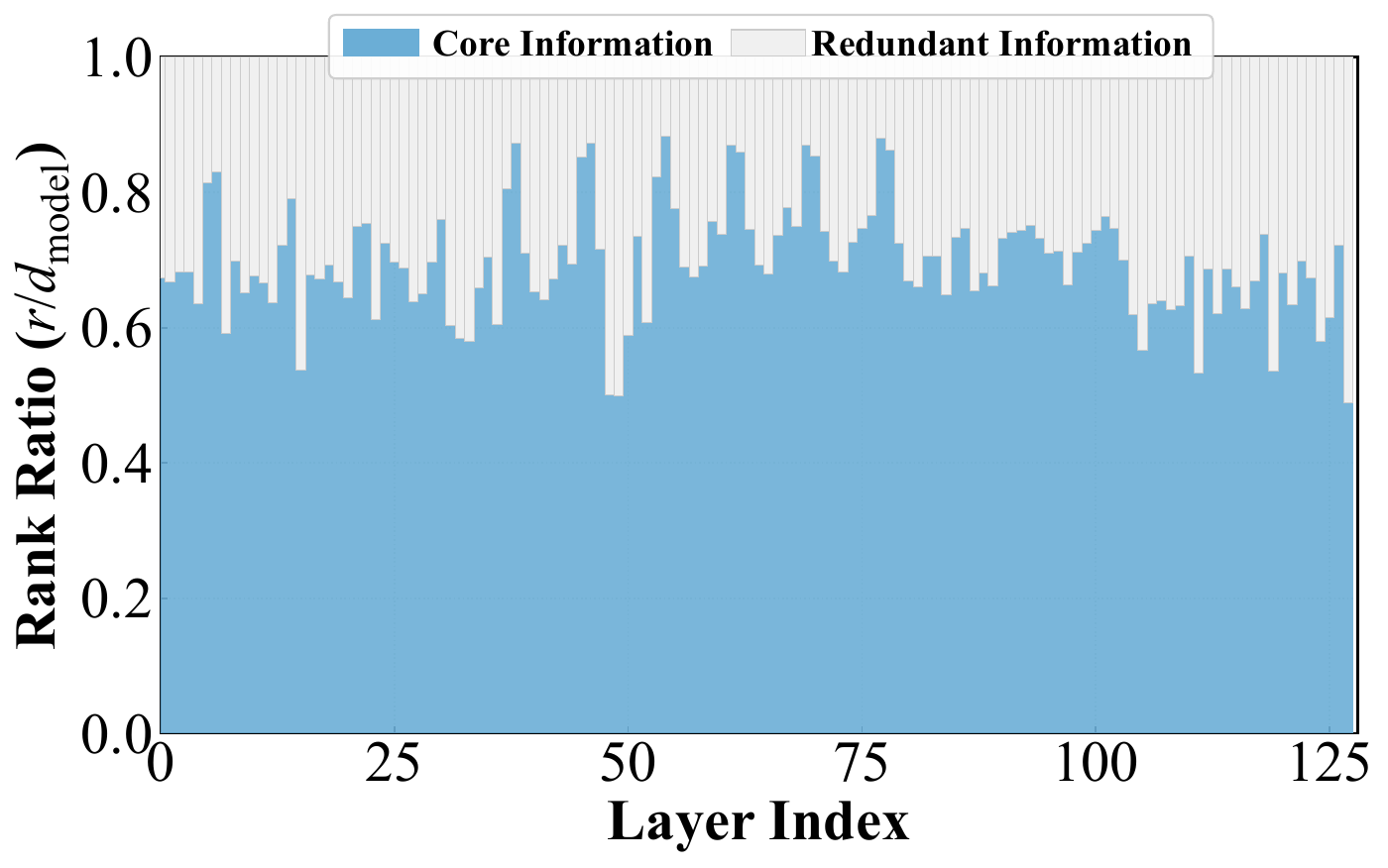}
        \caption{Effective dimensionality analysis}
        \label{fig:motivation_b}
    \end{subfigure}
    \hfill
    \begin{subfigure}[b]{0.42\linewidth}
        \centering
\includegraphics[width=\linewidth]{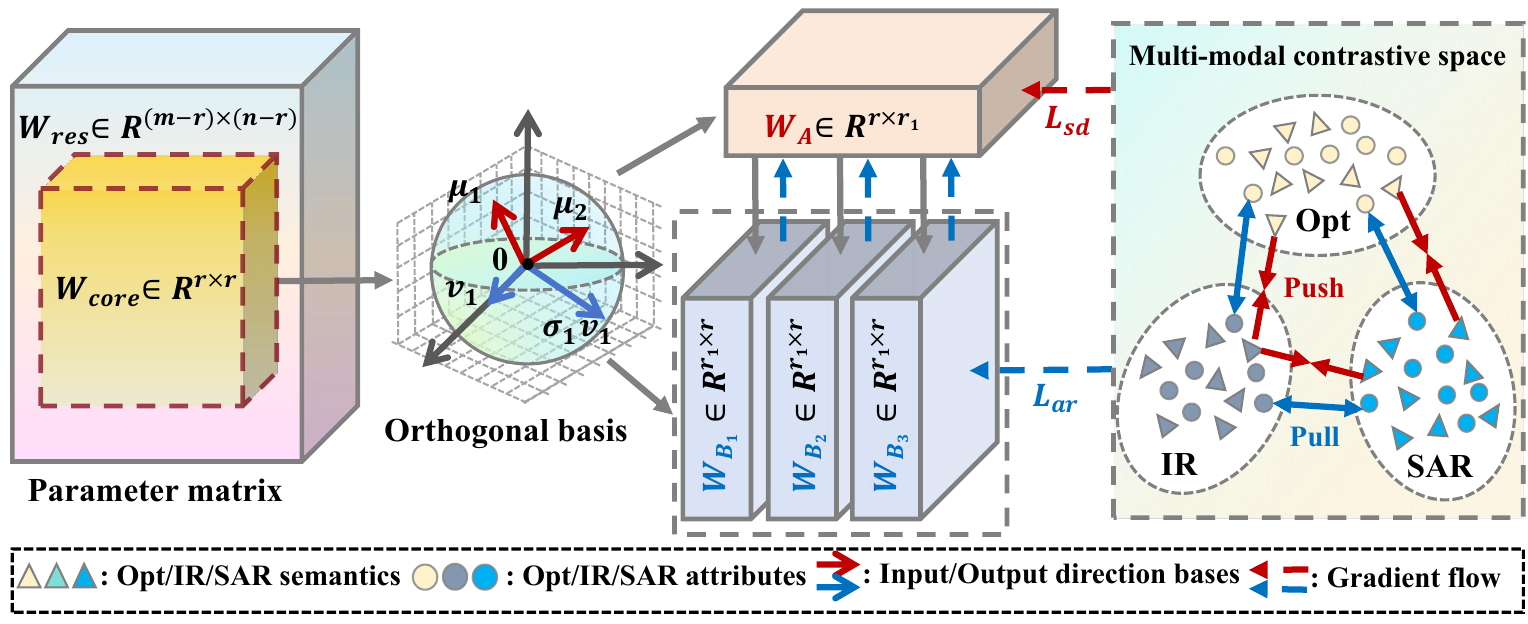}
        \caption{Motivation flowchart of our method}
        \label{fig:motivation_c}
    \end{subfigure}
    \caption{Statement of the motivation. (a) Similarity analysis of the parameter matrices $\mathbf{A}$ and $\mathbf{B}$ in the LoRA between the three modalities. (b) The layer-wise effective dimensionality analysis ($\frac{r}{d_{model}}$, $d_{model}$ represents the full dimension of the parameter matrix) of the UNet model, derived from the effective singular entropy computed of the parameter matrix. (c) In every network layer, the pre-trained parameter matrix $\mathbf{W}$ is decomposed into the core parameter matrix $\mathbf{W_{core}}$ and the redundant parameter matrix $\mathbf{W_{res}}$ based on its singular value distribution. Leveraging $\mathbf{W_{core}}$ as a fixed orthogonal basis, the parameter matrix $\mathbf{A}$ is optimized to disentangle invariant semantics across modalities via semantic disentanglement loss $\mathcal{L}_{sd}$, while different parameter matrices $\mathbf{B}_i$ ($i \in \{1, 2, 3\}$) are adapted to characterize distinct modal attributes via attribute reconstruction loss $\mathcal{L}_{ar}$.}
    \label{motivation} 
    \vspace{-0.5cm} 
\end{figure*}
\indent Specifically, CPD anchors the adaptation process within the core parameter space in all UNet layers of the SD
model, which is dynamically extracted from the pre-trained parameters based on the effective entropy of their singular values distribution of each layer. We next establish a disentangled optimization strategy (DOS) that is guided by the orthogonal basis derived from the core parameter subspace. Constrained by the input-side orthogonal basis, the parameter matrix $\mathbf{A}$ can adaptively extract invariant semantics via a multi-modal contrastive semantic disentanglement loss $\mathcal{L}_{sd}$ under the textual prompts. Meanwhile, the parameter matrices $\mathbf{B}$ are steered by the output-side orthogonal basis to model distinct modal attributes, supervised by an attribute reconstruction loss $\mathcal{L}_{ar}$. Furthermore, to achieve structural alignment between generated multimodal images, we further devise a query-key structure transfer (QKST) mechanism, which leverages the  self-attention map of the anchor modality to guide the target modality in aligning its spatial dependencies with the anchor. This mechanism enforces joint modeling of multimodal diffusion sampling trajectories in the structural attention space, ensuring structurally aligned multimodal generation while preserving each modality’s unique attributes. In these ways, the generated multimodal images exhibit high-fidelity OPT contrast, clear IR details, and rich SAR textures, and maintain cross-modal structural alignment (as shown in Fig. \ref{fig1s}). \\
\indent In summary, our main contributions are as follows:
\begin{itemize}
		\item Existing remote sensing image generation methods are confined to single-modality synthesis, failing to harness complementary multimodal information critical for accurate earth observation. We propose a text-to-multimodal remote sensing image generation task, which enables generating semantically consistent multimodal images from a single text prompt, thereby facilitating robust and comprehensive earth observation under diverse environmental conditions.   
		\item We propose a CPD module that leverages the functional dichotomy of LoRA adapters, performing parameter-level disentanglement within the orthogonal core subspace. By the devised disentangled optimization strategy, the parameter matrix $\mathbf{A}$ of the CPD module is contrastively optimized to extract invariant semantics, while the parameter matrices $\mathbf{B}_i$ are adapted to different modality attributes, thereby enabling high-fidelity multimodal image generation.
		\item We devise a QKST mechanism that injects structural correlation priors from an anchor modality into the target modalities during inference, guaranteeing spatial structure consistency across the generated multi-modal remote sensing images.
	\end{itemize}
\section{ RELATED WORK}
\subsection{Remote Sensing Image Generation} 
Remote sensing image generation differs significantly from general image synthesis due to the domain's unique overhead perspective and large-scale geographic context. Txt2Img-MHN \cite{xu2023txt2img} is the first work that employs modern Hopfield networks \cite{ramsauer2020hopfield} to achieve text-to-remote-sensing-image generation, laying the theoretical and practical foundation for the development of remote sensing image generation tasks. Subsequently, DiffusionSat \cite{khanna2024diffusionsat}, CRS-Diff \cite{10663449}, and GeoSynth \cite{sastry2024geosynth} adopt the ControlNet model \cite{zhang2023adding} to inject metadata (such as layout maps, depth maps, as well as weather and time conditions) into the baseline SD model \cite{rombach2022high}, guiding controllable remote sensing image generation and improving its generation quality. Text2earth \cite{10988859} proposes a global-scale remote sensing dataset and combines it with foundational generative models to achieve multi-resolution, unbounded text-to-remote-sensing-image synthesis. CC-Diff++ \cite{11187367} utilizes an advanced masked attention mechanism to model the interaction between foreground and background features during feature extraction. OTD-GAN \cite{11126950} designs a text decoupling module that enhances the focus on global and local targets, achieving precise object layout in remote sensing image generation. \\
\indent However, the aforementioned studies are limited to single-modal remote sensing image generation. In practice, different modalities of remote sensing images (e.g., OPT, IR, and SAR images) provide rich complementary information, especially under adverse conditions, which is critical for more comprehensive and reliable image synthesis. To this end, this paper proposes a text-to-multimodal remote sensing image generation task to fill the research gap.
\begin{figure*}[t]		
\setlength{\belowcaptionskip}{-0.5cm} \includegraphics[width=\textwidth]{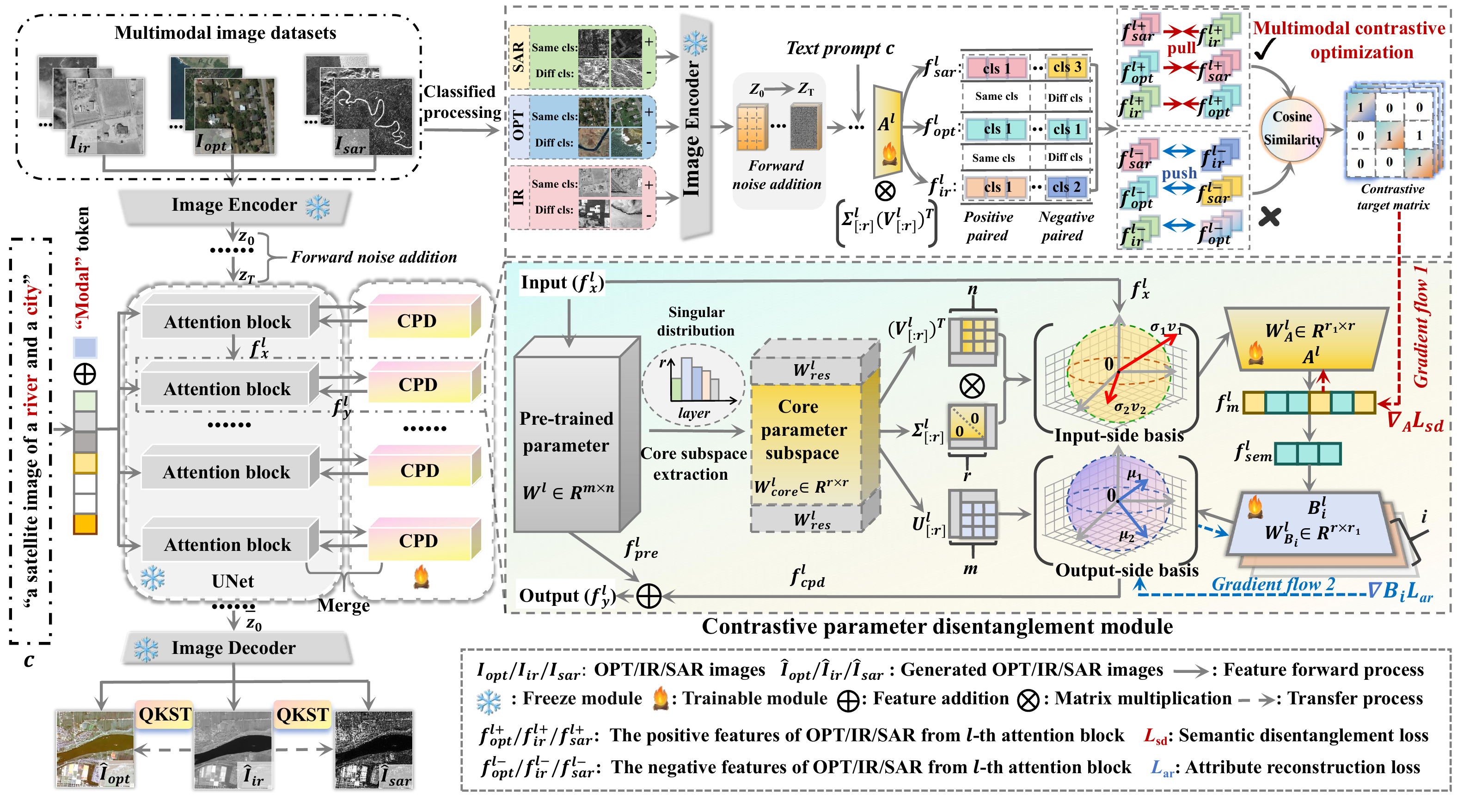}
		\centering 
		\caption{The training framework of our proposed CPD module. The pre-trained parameters are first decomposed into a core parameter subspace to derive orthogonal bases. Guided by these bases, the adaptation process is factorized into two stages: the parameter matrix $\mathbf{A}$ is constrained by the input-side basis to capture cross-modal invariant semantics via contrastive optimization, while the parameter matrix $\mathbf{B}$ consists of $i$ branches to reconstruct modality attributes under the output-side basis guidance via attribute reconstruction loss. Finally, the QKST mechanism is designed during inference to ensure spatial structural alignment across the generated multimodal images.}
		\label{fig4}
	\end{figure*}
\subsection{Text-to-Image Generation}
Text-to-image generation aims to synthesize visually faithful images from free-form text descriptions, and recent diffusion models have become the dominant paradigm due to their superior generation quality and sample diversity. Denoising diffusion probabilistic models \cite{ho2020denoising} establish the theoretical foundation by modeling image generation as a sequential denoising process. The denoising diffusion implicit models (DDIM) \cite{song2020denoising} improves the speed of diffusion denoising through deterministic sampling while ensuring the fidelity of the generated quality. To mitigate the enormous computational cost associated with pixel-space diffusion, the SD model \cite{rombach2022high} is proposed, which performs the diffusion process in a compressed latent space that balances computational efficiency and perceptual quality effectively. Other notable models, such as DALL-E2 \cite{ramesh2022hierarchical} and Imagen \cite{saharia2022photorealistic}, further demonstrate the scalability of diffusion models by integrating large-scale pre-trained text encoders (e.g., CLIP \cite{radford2021learning}), enabling the generation of photorealistic images with unprecedented semantic alignment. With the success of large foundational models, research focus has shifted toward domain-specific improvements to address unique practical challenges \cite{duan2025unic,liu2025alignguard,11391567,zhang2025cspanet}.\\
\indent Consequently, we adopt the SD model \cite{rombach2022high} as the generative backbone, capitalizing on its unparalleled proficiency in text-to-image synthesis and precise semantic alignment. Furthermore, its extensible architecture provides an ideal substrate for embedding our proposed adapter, thereby enabling text-to-multimodal image generation.
\subsection{Disentangled Representation Learning}
Disentangling content and style for controllable generation is a longstanding goal in computer vision. Conventional methods typically learn explicit latent representations and minimize mutual information to enforce independence between content and style factors. For example, Shan et al.\cite{shan2024learning} propose a dual-branch network that decouples content and distortion features in point cloud evaluation via mutual information minimization. Li et al. \cite{li2024learning} design a progressive decoupling framework using explicit content and style encoders to separate identity from attributes. Wang et al. \cite{wang2024disentangled} provide a group theory-based theoretical framework to formalize disentanglement in self-supervised learning. On the other hand, parameter-efficient fine-tuning methods-LoRA \cite{hu2022lora} is widely adopted for more efficient disentanglement. Recent works attempt to use multiple LoRAs to learn distinct concepts: ZipLoRA \cite{shah2024ziplora} proposes merging independently trained content and style LoRAs by optimizing their orthogonality to reduce interference, while QR-LoRA \cite{yang2025qr} leverages orthogonal mathematical properties via QR decomposition on diffusion weights to naturally decouple content and style during optimization. \\
\indent In contrast to these paradigms, our work anchors optimization in the parameter space and leverages the inherent functional dichotomy of LoRA adapters. By explicitly encouraging the matrix $\mathbf{A}$ to capture cross-modal invariant semantics and matrix $\mathbf{B}$ to model modality attributes, we achieve parameter-level disentanglement for multimodal remote sensing image generation from a single text prompt.
\section{PROPOSED METHOD}
In this paper, we propose a text-to-multimodal remote sensing image generation task to fill the research gap of current remote sensing image generation. To
address the challenge, we propose a CPD module that anchors the adaptation process within the core parameter subspace of the UNet model and leverages the functional dichotomy of LoRA adapters to achieve parameter-level disentanglement. With our devised DOS, the CPD's parameter matrix $\mathbf{A}$ is contrastively optimized to extract invariant semantics, while different $\mathbf{B}_i$ adapt to modality attributes for multimodal image generation. Furthermore, we devise a QKST mechanism to achieve structural alignment between multimodal generated images. We introduce the specific details below.\subsection{Overview}
 As illustrated in Fig. \ref{fig4}, we employ the SDv1.5 model \cite{rombach2022high} as the generative backbone. Specifically, multimodal images $\mathbf{I} \in \{\mathbf{I}_{opt}, \mathbf{I}_{ir}, \mathbf{I}_{sar}\}$ are first compressed into a low-dimensional latent space $\mathbf{z}_0 = \mathcal{E}(\mathbf{I})$ using a pretrained variational autoencoder \cite{kingma2013auto}. The diffusion process is modeled as a Markov chain consisting of forward noise addition and reverse denoising. This forward process progressively corrupts the initial latent feature $\mathbf{z}_0$ by injecting Gaussian noise $\mathbf{\epsilon}$ controlled by a variance schedule $\beta_t$. At any timestep $t$, the noisy latent state $\mathbf{z}_t$ is formulated as:
\begin{align}
    q(\mathbf{z}_t|\mathbf{z}_{t-1}) &:= \mathcal{N}(\mathbf{z}_t; \sqrt{1-\beta_t}\mathbf{z}_{t-1}, \beta_t\mathbb{I}), \\
    \mathbf{z}_t &= \sqrt{\bar{\alpha}_t}\mathbf{z}_0 + \sqrt{1-\bar{\alpha}_t}\mathbf{\epsilon}, \quad \mathbf{\epsilon} \sim \mathcal{N}(0, \mathbb{I}),
\end{align}
where $\bar{\alpha}_t = \prod_{i=1}^t \alpha_i$, $\alpha_t = 1 - \beta_t$, and $\mathbb{I}$ denotes the identity matrix, $\mathcal{N}$ denotes the Gaussian distribution.\\
\indent For the reverse process, we aim to iteratively reconstruct the clean latent $\mathbf{z}_0$ from the Gaussian latent $\mathbf{z}_T$ via the inverse chain. Within the denoising UNet $\epsilon_\theta$ (parameterized by $\theta$), we merge the CPD module (parameterized by $\Theta_{cpd}$) into all its attention layers to modulate feature outputs. The denoising UNet performs the state transition from $\mathbf{z}_t$ to $\mathbf{z}_{t-1}$ by estimating the noise residual at each timestep:
\begin{equation}
    \mathbf{z}_{t-1} = \frac{1}{\sqrt{\alpha_t}} \left( \mathbf{z}_t - \frac{1-\alpha_t}{\sqrt{1-\bar{\alpha}_t}} \epsilon_{\theta \uplus\Theta_{cpd}}(\mathbf{z}_t, t, c) \right) + \sigma_t \mathbf{\xi},
    \label{eq:reverse_process}
\end{equation}
where $\mathbf{\xi} \sim \mathcal{N}(0, \mathbb{I})$ is the stochastic noise term and $c$ denotes the textual condition, the $\sigma_t$ noise standard deviation during the $t$-th step of reverse denoising, and $\uplus$ denotes the parameter
merge operation via addition.\\
\indent Following the reverse denoising steps, the denoised clean latents $\mathbf{\bar{z}}_0$ are projected back to the pixel space via the frozen image decoder, yielding the final generated multimodal images $\mathbf{\hat{I}} = \mathcal{D}(\mathbf{\bar{z}}_0)$ ($\mathbf{\hat{I}} \in \{\mathbf{\hat{I}}_{opt}, \mathbf{\hat{I}}_{ir}, \mathbf{\hat{I}}_{sar}\}$). To further ensure structural alignment across the generated multimodal images, we devise a QKST mechanism for the inference stage. Below, we elaborate on the structure design and optimization strategy of the proposed CPD module, as well as the implementation details of the QKST mechanism.
\subsection{Contrastive Parameter Disentanglement}
In this section, we elaborate on the structural design of the proposed CPD module. Taking the $l$-th attention block as an example (illustrated in Fig.~\ref{fig4}), let $\mathbf{W}^{l} \in \mathbb{R}^{m \times n}$ denote its parameter matrix. To distill the salient information while alleviating redundancy, we first perform singular value decomposition \cite{alter2000singular} on the parameter matrix $\mathbf{W}^{l}$, yielding three parameter matrices ($\mathbf{U}^{l}$, $\mathbf{\Sigma}^{l}$, and $\mathbf{V}^{l}$):
\begin{equation}
\begin{split}
\setlength{\abovedisplayskip}{5pt}   
\setlength{\belowdisplayskip}{5pt}
    \mathbf{W}^{l} &= \mathbf{U}^{l} \mathbf{\Sigma}^{l} (\mathbf{V}^{l})^\top = \sum_{k=1}^{K= \min(m,n)} \mathbf{u}_{l,k}\sigma_{l,k} \mathbf{v}_{l,k}^\top \\
    &= \mathbf{u}_{l,1}\sigma_{l,1}\mathbf{v}_{l,1}^\top + \mathbf{u}_{l,2}\sigma_{l,2}\mathbf{v}_{l,2}^\top + \cdots + \mathbf{u}_{l,K}\sigma_{l,K} \mathbf{v}_{l,K}^\top,
\end{split}
\label{eq1}
\end{equation}
where $\mathbf{U}^l, \mathbf{V}^l$ are orthogonal matrices, and $\mathbf{u}_{l,k}, \mathbf{v}_{l,k}$ denote their $k$-th column singular vectors, respectively. The diagonal matrix $\mathbf{\Sigma}^{l}$ contains the singular values sorted in descending order, i.e., $\sigma_{l,1} \ge \sigma_{l,2} \ge \dots \ge \sigma_{l,K} \ge 0$. \\
\indent Subsequently, we normalize the singular values $\mathbf{\sigma}_{l}$ to construct a probability distribution $P_l$ that quantifies the information density of each attention block, formulated as:
\begin{equation}
\setlength{\abovedisplayskip}{3pt}   
\setlength{\belowdisplayskip}{3pt}
P_{l,k} = \frac{\sigma_{l,k}^2}{\|\mathbf{W}^l\|_F^2}, \quad \text{s.t.} \quad P_l \in \left\{ \mathbf{p} \in \mathbb{R}^K_{\ge 0} \;\middle|\; \sum_{k=1}^K p_k = 1 \right\},
\end{equation}
where $P_{l,k}$ denotes the normalized energy ratio of the $k$-th singular value in the $l$-th attention block, which characterizes its relative contribution to the total information energy of $\mathbf{W}^l$, and $\|.\|_F$ denotes the Frobenius norm calculation. 

Building on the probability distribution $P_l$ of singular values, we adopt information entropy as a mathematical metric to characterize information density within each attention block. The higher entropy value indicates richer useful information, while the lower value suggests greater parameter redundancy. This is formally defined as the following: 
\begin{equation}
\setlength{\abovedisplayskip}{3pt}   
\setlength{\belowdisplayskip}{3pt}
    H(\mathbf{W}^l) = - \sum_{k=1}^K P_{l,k} \log P_{l,k}.
    \label{eq:spectral_entropy}
\end{equation}
We then derive the effective rank of the attention block by using the rounding  ($\lceil.\rceil$) to the exponential of the entropy:\\
\begin{equation}
    r_{eff}(\mathbf{W}^l) = \lceil \exp(H(\mathbf{W}^l)) \rceil.
    \label{eq:effective_rank}
\end{equation}
\indent Through the above transformation, we extract a rank-$r$ parameter subspace from the original matrix $\mathbf{W}^l$ as the core parameter matrix $\mathbf{W}^{l}_{core}$ $\in$ $\mathbb{R}^{m \times n}$, where the effective rank $r$=$r_{eff}$ is adaptively quantified for each attention block. This core parameter matrix is defined as follows:
\begin{equation}
\begin{split}
\setlength{\abovedisplayskip}{3pt}   
\setlength{\belowdisplayskip}{3pt}
    \mathbf{W}^{l}_{core} &= \mathbf{U}^{l}_{[:r]} \mathbf{\Sigma}^{l}_{[:r]} (\mathbf{V}^{l}_{[:r]})^\top = \sum_{k=1}^{r} \mathbf{u}_{l,k} \sigma_{l,k} \mathbf{v}_{l,k}^\top \\
    &= \mathbf{u}_{l,1} \sigma_{l,1} \mathbf{v}_{l,1}^\top + \mathbf{u}_{l,2} \sigma_{l,2} \mathbf{v}_{l,2}^\top + \cdots + \mathbf{u}_{l,r} \sigma_{l,r} \mathbf{v}_{l,r}^\top,
\end{split}
\label{eq6}
\end{equation}
where $[:]$ denotes the matrix slicing operation.\\
\indent We next factorize this core subspace into two structurally complementary bases. As illustrated on the right of Fig. \ref{fig4}, the weighted right singular components $\mathbf{\Sigma}^{l}_{[:r]} (\mathbf{V}^{l}_{[:r]})^\top$ serve as the \textbf{input-side basis}  $\mathcal{P}_{in}^l$, while the left singular components $\mathbf{U}^{l}_{[:r]}$ serve as the \textbf{output-side basis}  $\mathcal{P}_{out}^l$. These structural priors ensure mutually independent directions. We then instantiate the LoRA adapter within this constrained space. Specifically, the parameter matrix $\mathbf{A}^l$ ($\mathbf{W}^l_A \in \mathbb{R}^{ r_1 \times r}$, $r_1$ denotes the rank dimension, $r_1$ $\le$ $r$) is instantiated based on the input-side basis to act as the semantic disentangler, and the parameter matrix $\mathbf{B}^l$ ($\mathbf{W}^l_B \in \mathbb{R}^{r \times  r_1}$) is instantiated based on the output-side basis to serve as the attribute adapter. To accommodate multimodal attributes without mutual interference, we expand three parameter matrices $\mathbf{B}_i$ ($i \in \{1, 2, 3\}$), thus the parameter structure of the CPD module is formulated as:
\begin{equation}
\begin{split}
\setlength{\abovedisplayskip}{3pt}   
\setlength{\belowdisplayskip}{3pt}
\Delta \mathbf{W}^l_{\text{CPD}} &= (\mathbf{U}_{[:r]}^l \mathbf{B}_i^l)(\mathbf{A}^l \mathbf{\Sigma}_{[:r]}^l (\mathbf{V}_{[:r]}^l)^T) \\
&= \underbrace{(\big[ \mathbf{u}_{l,1}, \dots, \mathbf{u}_{l,r} \big] \mathbf{B}_i^l)}_{\text{attribute adapter}} \underbrace{((\mathbf{A}^l)
\begin{bmatrix}
\sigma_{l,1} \mathbf{v}_{l,1}^T \\
\vdots \\
\sigma_{l,r} \mathbf{v}_{l,r}^T
\end{bmatrix})}_{\text{semantic disentangler}} \\
\end{split}
\label{eq:cpd_update_expanded}
\end{equation}
where $\Delta \mathbf{W}_{\text{CPD}}^l$ denotes the parameter structure of the CPD module merged into the $l$-th attention block.\\
\indent Finally, with the integration of the CPD module and the $l$-th attention block, the output feature $\mathbf{f}^l_{y}$ is modulated as:
\begin{equation}\label{eq:final_feature}
\setlength{\abovedisplayskip}{3pt}   
\setlength{\belowdisplayskip}{3pt}
\mathbf{f}^l_y = \mathbf{f}^l_{pre} + \mathbf{f}^l_{cpd}, \quad \text{where} \begin{cases}
    \mathbf{f}^l_{pre} =  \mathbf{W}^l \mathbf{f}_{x}^l \\
    \mathbf{f}^l_{cpd} = \Delta \mathbf{W}_{\text{CPD}}^l  \mathbf{f}_{x}^l
\end{cases}
\end{equation}
where $\mathbf{f}_{x}^l$ denotes the input feature of the 
$l$-th attention block.\\
\begin{algorithm}[t]
\caption{Structure and optimization of the CPD}
\label{alg:cpd_training}
\begin{algorithmic}[1]
\REQUIRE Pre-trained parameter matrix $\mathbf{W}$, multi-modal dataset $\mathcal{D} = \{\mathbf{I}_{opt}, \mathbf{I}_{ir}, \mathbf{I}_{sar},$c$\}$, learning rate: $\eta_1$, $\eta_2$;
\ENSURE Optimized parameter matrices $\mathbf{A}$ and $\{\mathbf{B}_i\}_{i=1}^{i=3}$.
\STATE \textbf{// Structure Construction}
\FOR{each attention layer $l \in \{1, \dots, L\}$}
    \STATE Perform SVD: $\mathbf{U}^l, \mathbf{\Sigma}^l, (\mathbf{V}^l)^T \leftarrow \text{SVD}(\mathbf{W}^l)$
    \STATE Determine effective rank: $r \leftarrow \exp(H(\mathbf{W}^l))$ \COMMENT{Eq.~\eqref{eq:effective_rank}}
    \STATE Extract orthogonal bases:
    \STATE \quad Input-side: $\mathcal{P}_{in}^l \leftarrow \mathbf{\Sigma}^l[:r] (\mathbf{V}^l)^T[:r, :]$
    \STATE \quad Output-side: $\mathcal{P}_{out}^l \leftarrow \mathbf{U}^l[:r]$
    \STATE Initialize parameter matrices: $\mathbf{A}^l, \mathbf{B}_{i}^l$
\ENDFOR
\STATE \textbf{// Phase I - Semantic Disentanglement}
\WHILE{not converged}
    \FOR{each attention layer $l \in \{1, \dots, L\}$}
        \STATE \textbf{Freeze:} $\mathcal{P}_{in}^l, \mathcal{P}_{out}^l, \mathbf{B}_i^l$; \textbf{Trainable:} $\mathbf{A}^l$
        \STATE Sample multi-modal batch $\{\mathbf{I}_u, \mathbf{I}_v, c\}$ from $\mathcal{D}$
        \STATE Extract invariant semantics: $\mathbf{f}^l_{sem} \leftarrow \mathbf{A}^l \cdot \mathcal{P}_{in}^l \mathbf{f}^l_x$
        \STATE Compute disentanglement loss: $\mathcal{L}_{sd}$ \COMMENT{Eq.~\eqref{eq:total_sd_loss}}
        \STATE Update $\mathbf{A}^l \leftarrow \mathbf{A}^l - \eta_1 \nabla \mathcal{L}_{sd}$ 
    \ENDFOR  
\ENDWHILE
\STATE \textbf{// Phase II - Attribute Adaptation}
\WHILE{not converged}
    \FOR{each modality $i \in \{\text{1, 2, 3}\}$}
        \FOR{each attention layer $l \in \{1, \dots, L\}$}
            \STATE \textbf{Freeze:} $\mathcal{P}_{in}^l, \mathcal{P}_{out}^l, \mathbf{A}^l$; \textbf{Trainable:} $\mathbf{B}_i^l$
            \STATE Sample batch $(\mathbf{z}_t, \bar{c})$ for modality $i$
            \STATE Compute reconstruction loss: $\mathcal{L}_{ar}$ \COMMENT{Eq.~\eqref{eq:attr_loss}}
            \STATE Update $\mathbf{B}_i^l \leftarrow \mathbf{B}_i^l - \eta_2 \nabla \mathcal{L}_{ar}$
        \ENDFOR  
    \ENDFOR  
\ENDWHILE  
\RETURN Parameters matrices $\{\{\mathbf{A}\}_{l=1}^L, \{\mathbf{B}_i\}_{l=1}^L\}$
\end{algorithmic}
\end{algorithm}
\textbf{\textit{Theoretical analysis of the CPD module:}}
\label{sec:theoretical_analysis}
We theoretically analyze the structural advantages of the CPD module from a linear algebraic perspective, focusing on redundancy isolation and structural disentanglement.\\
\indent \textit{Property 1: Redundancy Isolation.} 
Based on the aforementioned decomposition of $\mathbf{W}^l$, we split it into a core parameter matrix $\mathbf{W}_{core}^l$ and a redundant parameter matrix  $\mathbf{W}_{res}^l = \mathbf{U}_{[r:]}^l \Sigma_{[r:]}^l (\mathbf{V}_{[r:]}^l)^\top$. On the input side, the CPD module projects the input feature $\mathbf{f}_x^l$ exclusively via the input-side basis $\mathcal{P}_{in}^l$=$\mathbf{\Sigma}^{l}_{[:r]} (\mathbf{V}^{l}_{[:r]})^\top$. This operation inherently isolates the feature from the redundant parameter subspace $\text{span}(\mathbf{V}_{[r:]}^l)$ due to the strict orthogonality condition:
\begin{equation}
\mathcal{P}_{in}^l\mathbf{V}_{[r:]}^l=\mathbf{\Sigma}^{l}_{[:r]} \underbrace{(\mathbf{V}^{l}_{[:r]})^\top\mathbf{V}_{[r:]}^l}_{\text{0}} = \mathbf{0}.
\end{equation}
Symmetrically, the output feature $\mathbf{f}_{cpd}^l$ is constrained within the core output-side subspace $\text{span}(\mathbf{U}_{[:r]}^l)$, satisfying:
\begin{equation}
\setlength{\abovedisplayskip}{3pt}   
\setlength{\belowdisplayskip}{3pt}   
    \mathbf{f}_{cpd}^l \in \text{span}(\mathbf{U}_{[:r]}^l) \implies (\mathbf{U}_{[r:]}^l)^\top \mathbf{f}_{cpd}^l = \mathbf{0}.
\end{equation}
This dual-side orthogonality mathematically isolates the CPD feature flow from the redundant space $\mathbf{W}_{res}^l$, thereby
shielding optimization from task-irrelevant noise and ensuring the model focuses on only useful information.\\
\indent \textit{Property 2: Structural Disentanglement.} CPD achieves a two-level disentanglement by exploiting the orthogonality of the input-output side bases. (i) \textit{Semantic-Attribute Separation}: The invariant semantics are extracted via the input-side basis $\mathcal{P}_{in}^l$, while modal attributes are injected via the output-side basis $\mathcal{P}_{out}^l$. Due to $\operatorname{span}\left(\mathcal{P}_{in}^l\right) \neq \operatorname{span}\left(\mathcal{P}_{out}^l\right)$, this naturally ensures that semantic and attribute updating operate on distinct orthogonal spaces.
(ii) \textit{Inter-attribute Isolation}: Multiple parallel parameter matrices $\mathbf{B}^l_i$ are independently modulated. For distinct matrices $i \neq j$ and input feature $\mathbf{f}_x^l$, their mutual interference is characterized as:
\begin{equation}
\begin{aligned}
\label{13}
\setlength{\abovedisplayskip}{3pt}   
\setlength{\belowdisplayskip}{3pt}
    \langle \Delta \mathbf{f}^l_i, \Delta \mathbf{f}^l_j \rangle 
    &= (\mathbf{U}^l_{[:r]} \mathbf{B}^l_i \mathbf{A}^l \mathcal{P}_{in}^l \mathbf{f}_x^l)^{\top} (\mathbf{U}^l_{[:r]} \mathbf{B}^l_j \mathbf{A}^l \mathcal{P}_{in}^l \mathbf{f}_x^l) \\
    &= (\mathbf{f}_x^l\mathcal{P}_{in}^l\mathbf{A}^l\mathbf{B}^l_i)^{\top} \underbrace{(\mathbf{U}^l_{[:r]})^{\top} \mathbf{U}^l_{[:r]}}_{\mathbf{I}_{r \times r}} \mathbf{B}^l_j \mathbf{A}^l \mathcal{P}_{in}^l \mathbf{f}_x^l \\
    &= (\mathbf{f}_x^l)^{\top}(\mathcal{P}_{in}^l)^{\top} (\mathbf{A}^l)^{\top} \left[ (\mathbf{B}^l_i)^{\top} \mathbf{B}^l_j \right] \mathbf{A}^l\mathcal{P}_{in}^l \mathbf{f}_x^l.
\end{aligned}
\end{equation}
Due to the orthonormality of the core singular basis $\mathbf{U}^l_{[:r]}$, all high-dimensional coupling terms are eliminated in Eq.~\eqref{13}.
The only residual interaction between different modality branches is determined by the parameter correlation $(\mathbf{B}^l_i)^\top \mathbf{B}^l_j$.
Accordingly, independently optimizing the parameter matrices $\mathbf{B}_i$ suppresses this correlation, theoretically avoiding inter-attribute interference and further improving disentanglement performance.
\begin{figure}[t]
\includegraphics[width=0.49\textwidth]{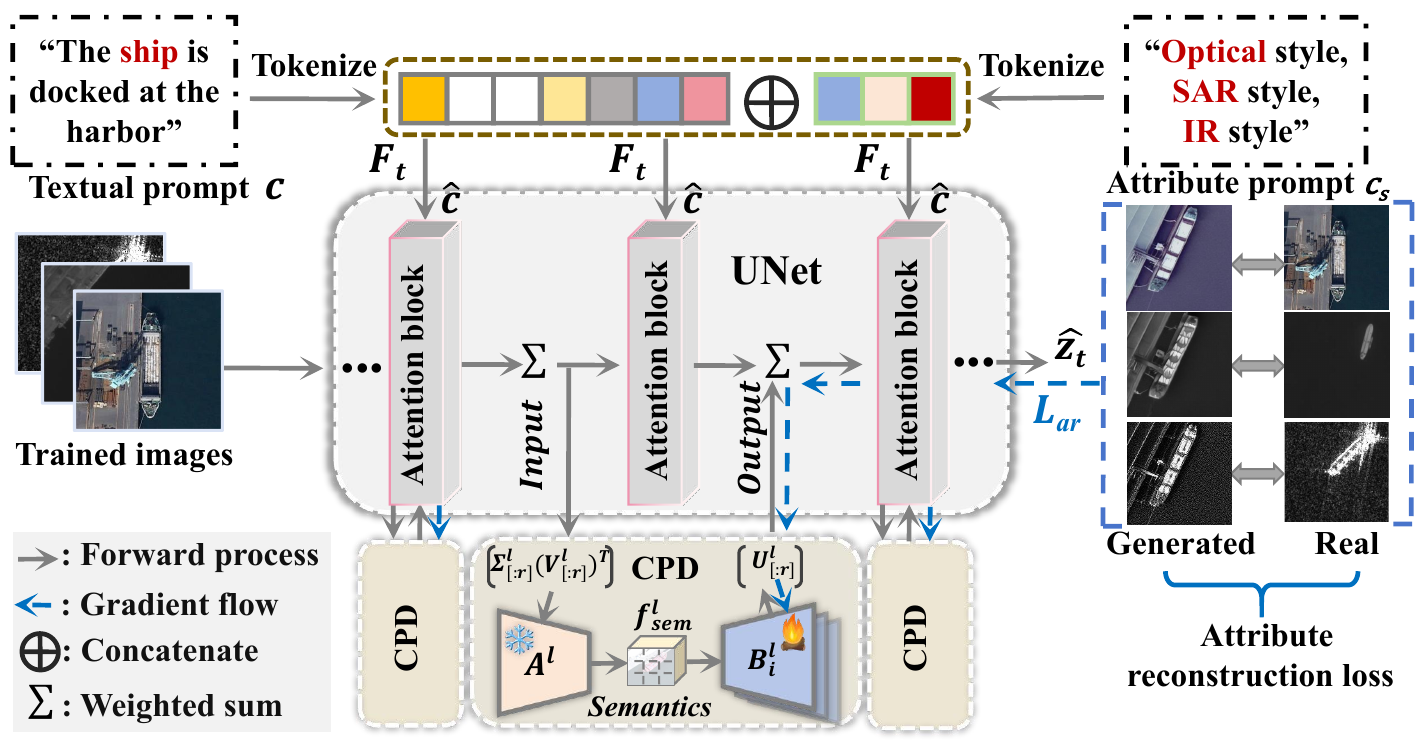}
		\centering \caption{Illustration of the attribute reconstruction loss. We concatenate the base text prompt $c$ with the attribute prompt $c_s$ to form the composite condition prompt $\hat{c}$ to optimize the different parameter matrices $\mathbf{B}_i$.}
		\label{fig4s}
        \vspace{-10pt}
	\end{figure} 
\subsection{Disentangled Optimization Strategy}
\label{dso}
Building upon the proposed CPD module, we build a dedicated DOS that optimizes the parameter matrix $\mathbf{A}$ with a semantic disentanglement loss ($\mathcal{L}_{sd}$) to extract invariant semantics, and optimizes the parameter matrices $\mathbf{B}_i$ with a attribute reconstruction loss ($\mathcal{L}_{ar}$) to adapt to distinct modal attributes. We provide a detailed introduction to DOS below.\\
\indent \textbf{Semantic disentanglement loss:} As a core component of our DOS, this loss is designed to regularize the parameter matrix $\mathbf{A}$ to capture invariant semantics across all modalities, which serves as the semantic anchor for multimodal generation. Our key insight is that features projected by $\mathbf{A}^l$ should converge to a consistent semantic space defined by the text prompt, regardless of the target modality (OPT, IR, or SAR). To operationalize this, within the reverse denoising process defined in Eq.~\eqref{eq:reverse_process}, we take the intermediate input feature $\mathbf{f}_x^l$ from the $l$-th attention block, and project it via the parameter matrix $\mathbf{A}^l$ to obtain invariant semantic features: 
\begin{align}
    \mathbf{f}_m^l =\left(\mathbf{A}^l \left( \mathbf{\Sigma}^{l}_{[:r]} (\mathbf{V}^{l}_{[:r]})^T\right)\right)\cdot  \mathbf{f}_x^l, \quad m \in \{opt, ir, sar\}, \label{eq:sem_feature}
\end{align}
where $\mathbf{f}_m^{l}$ denotes the semantic feature projected by $\mathbf{A}^l$ for modality $m$, as illustrated in the top-right of Fig. \ref{fig4}.

\indent Based on these projected semantic features, we construct multimodal contrastive constraints to pull semantically consistent features together and push dissimilar ones apart. Specifically, we form positive feature pairs ($\mathbf{f}_{opt}^{l+}$, $\mathbf{f}_{ir}^{l+}$, $\mathbf{f}_{sar}^{l+}$) from multimodal images sharing the same semantic label, and negative pairs ($\mathbf{f}_{opt}^{l-}$, $\mathbf{f}_{ir}^{l-}$, $\mathbf{f}_{sar}^{l-}$) from images of distinct semantic classes. The multimodal contrastive loss between arbitrary modalities $u, v \in \{opt, ir, sar\}$ is then defined as:
\begin{equation}
    \mathcal{L}^l_{c}(u, v, c) = -\sum_{i \in \mathcal{B}} \frac{1}{|P_i(v)|} \sum_{p \in P_i(v)} \log \frac{e^{\text{sim}(\mathbf{f}^{l+,i}_u, \mathbf{f}^{l+,p}_v,c) / \tau}}{\sum_{k \in \mathcal{B}} e^{\text{sim}(\mathbf{f}^{l+,i}_u, \mathbf{f}^{l,k}_v,c) / \tau}},
    \label{eq15}
\end{equation}
where $\mathcal{B}$ denotes the set of sample indices in the current batch. $i \in \mathcal{B}$ represents the anchor sample index in modality $u$, and its corresponding feature is denoted as $\mathbf{f}^{l+,i}_u$. $P_i(v)$ is the set of indices of all positive samples in modality $v$ that share the same semantic class as sample $i$. Accordingly, $\mathbf{f}^{l+,p}_v$ denotes the positive paired features ($p \in P_i(v)$) that are pulled closer to the anchor. $|P_i(v)|$ denotes the cardinality of the positive set. $k \in \mathcal{B}$ traverses all batch samples for normalization, with $k \notin P_i(v)$ as implicit negative pairs to enforce semantic discrimination. $\text{sim}(\cdot, \cdot, c)$ denotes the cosine similarity conditioned on the text prompt $c$, with temperature hyperparameter $\tau = 0.07$ by default.\\
\indent To achieve fundamentally semantic disentanglement, we extend this constraint $\mathcal{L}_{c}^l$ to all attention blocks across the full UNet hierarchy ($l \in \{1, \dots, L\}$). Consequently, the final semantic disentanglement loss $\mathcal{L}_{sd}$ is formulated as:
\begin{align}
\setlength{\abovedisplayskip}{5pt}   
\setlength{\belowdisplayskip}{5pt}
    \mathcal{L}_{sd} = \sum_{l=1}^{L} \left( \mathcal{L}_{c}^l(opt, ir,c)+ \mathcal{L}_{c}^l(opt, sar,c)+ \mathcal{L}_{c}^l(ir, sar,c) \right),
    \label{eq:total_sd_loss}
\end{align}
\begin{figure*}[t]
\includegraphics[width=0.95\textwidth]{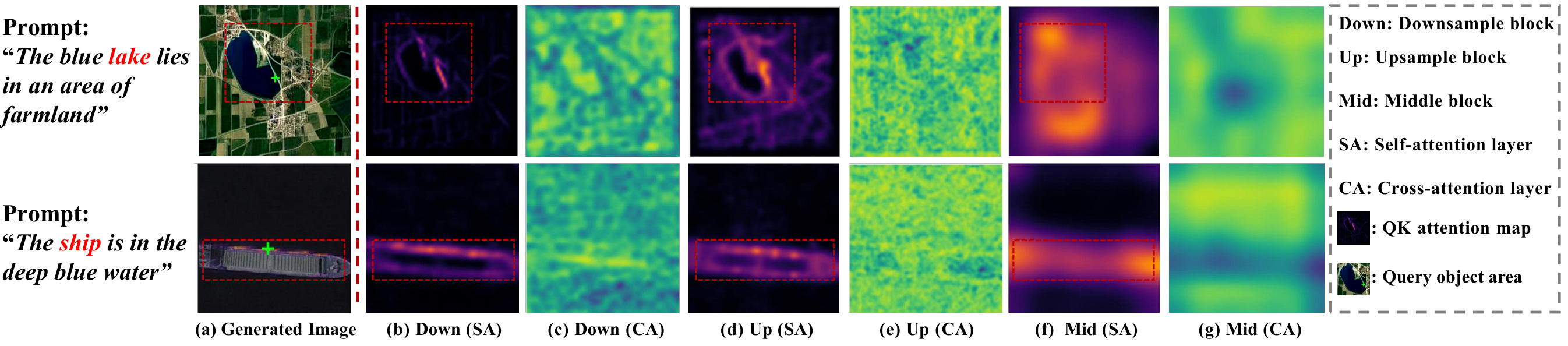}
\centering
		\caption{Visualizations of $QK$ attention activation maps across distinct UNet stages, including SA and CA heatmaps from the encoder (Down), middle (Mid), and decoder (Up) blocks.}
		\label{qkst}
        \vspace{-10pt}
	\end{figure*}  
    \begin{figure*}[t]
\includegraphics[width=0.95\textwidth]{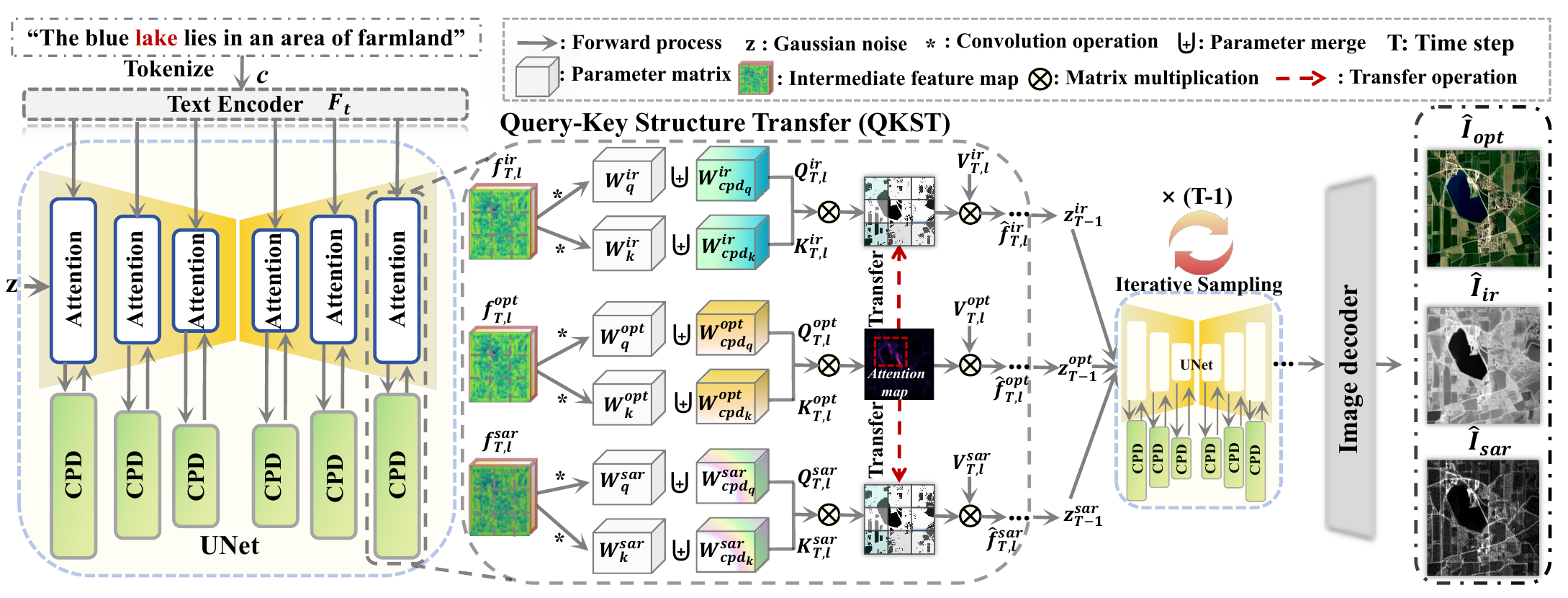}
		\centering \caption{Overview framework of the multimodal generation inference. Leveraging only a single text prompt $c$, the OPT branch provides structural guidance via the QKST mechanism, which transfers attention maps to IR and SAR branches for consistent spatial structural alignment.}
\label{fig:QKST_inference}
        \vspace{-10pt}
	\end{figure*} 
where $L$ denotes the total number of UNet attention blocks.\\
\indent \textbf{Attribute reconstruction loss:} 
As the second stage of our DOS, we aim to project the invariant semantics learned by
$\mathbf{A}$ onto modality-specific manifolds by optimizing the parameter matrices $\mathbf{B}_i$. To enable explicit modal adaptation, as illustrated in Fig. \ref{fig4s}, we first construct a composite condition prompt $\hat{c}$ by concatenating the base textual prompt $c$ with the modality-corresponding attribute prompt $c_s$:
\begin{equation}
   \hat{c} = F_{t}((Tokenize(c)) \oplus (Tokenize(c_s))),
   \label{eq:composite_prompt}
\end{equation}
where $F_{t}$ denotes the pre-trained text encoder and $Tokenize(\cdot)$ denotes the tokenize operation, $\oplus$ denotes the concatenate operation.\\
\indent We then feed the composite condition prompt $\hat{c}$ into the UNet model to guide the diffusion denoising process. To preserve the invariant semantics, we freeze the parameter matrix $\mathbf{A}$, such that the text-anchored semantic feature $\mathbf{f}^l_{sem}$ remains unchanged. We optimize independently each $\mathbf{B}_i$ end-to-end under the guidance of $\hat{c}$ to exclusively model the corresponding modal attributes, and cast its optimization as a diffusion denoising objective since its output features are directly integrated into the UNet pipeline. Accordingly, the attribute reconstruction loss $\mathcal{L}_{ar}$ is defined as:
\begin{equation}
    \mathcal{L}_{ar} = \min_{B_i} \mathbb{E}_{\mathbf{z}_0, \hat{c}, t, \mathbf{\epsilon}} \left[ \| \mathbf{\epsilon} - \epsilon_{\theta \uplus \Theta_{cpd}}(\mathbf{\hat{z}}_t, t, \hat{c}) \|_2^2 \right],
    \label{eq:attr_loss}
\end{equation}
where $\mathbf{\hat{z}}_t$ denotes the feature predicted by the UNet model at the timestep $t$ following modulation by our CPD module.\\
\indent Note that the composite condition prompt $\hat{c}$ is used only in training as a regularization to guide the optimization of $\mathbf{B}_i$. After training, modality identifiers $c_s$ are discarded, and multimodal inference only requires a single text prompt $c$. Through this disentangled optimization strategy, our CPD module can accurately map invariant semantics under the guidance of the textual prompt while adapting to different modal attributes. We provide the structure and optimization details of the CPD module in Algorithm \ref{alg:cpd_training}.
\subsection{Query-Key Structural Transfer}\label{QKST}
Stochastic sampling trajectories in the diffusion process cause misalignment of cross-modal structures, which impairs practical utility in downstream
tasks such as image fusion, where strict cross-modal spatial correspondence is essential. To diagnose the cause, we analyze the attention behavior within the UNet. As shown in Fig.~\ref{qkst}, we observe that cross-attention (CA) layers activate at text-specified object locations (e.g., lake, ship), encoding coarse-grained semantic alignment via low-frequency responses. Conversely, self-attention (SA) layers exhibit broader spatial activation, capturing structural relationships and topological boundaries. This distinction confirms that SA layers dominate spatial structure, justifying joint modeling of multimodal sampling trajectories for alignment.\\
\indent To this end, we design the QKST mechanism to embed all the SA layers of the UNet model during the inference process. Fig. \ref{fig:QKST_inference} illustrates the inference framework with QKST for multimodal generation. Specifically, let $\mathbf{z}_t^m$ denote the latent feature of modality $m \in \{opt, ir, sar\}$ at diffusion step $t \in [0, T]$. Given the text prompt $c$, let $\mathbf{f}_{t,l}^m$ denotes the intermediate feature input to the $l$-th SA block of the UNet model $\epsilon_\theta(\mathbf{z}_t^m, t, c)$ (illustrated as the $T$-th step $\mathbf{f}_{T,l}^m$ in Fig. \ref{fig:QKST_inference}). The pre-trained parameters are modulated by our CPD module via a parameter merge operation ($\uplus$). Using the convolution operation ($*$) as defined in the network, the modulated query ($\mathbf{Q}$) and key ($\mathbf{K}$) matrices are defined as:
\begin{align}
\mathbf{Q}_{t,l}^m &= (\mathbf{W}_q^m \uplus \mathbf{W}^{m}_{cpd_q}) * \mathbf{f}_{t,l}^m, \label{eq:Q_update} \\
\mathbf{K}_{t,l}^m &= (\mathbf{W}_k^m \uplus  \mathbf{W}_{cpd_k}^m) * \mathbf{f}_{t,l}^m, \label{eq:K_update}
\end{align}
where $\mathbf{W}_q^m$ and $\mathbf{W}_k^m$ denote the pre-trained projection parameters in the UNet. The $\mathbf{W}_{cpd_q}^m$ and $\mathbf{W}_{cpd_k}^m$ denote the query and key projection parameters from the CPD module. \\
\indent To enforce cross-modal structural consistency at each diffusion step, we adopt the OPT branch as the geometric anchor, which provides reliable structural priors for other modalities. We compute the attention map $\mathcal{A}_{t,l}^{opt}$ from the OPT branch via Q-K matrix multiplication:
\begin{equation}
\label{19}
\mathcal{A}_{t,l}^{opt} = \text{Softmax}\left( \frac{\mathbf{Q}_{t,l}^{opt} \otimes (\mathbf{K}_{t,l}^{opt})^\top}{\sqrt{d}} \right).
\end{equation}
where $d$ denotes the scaling dimension of the Q-K matrices, and $\otimes$ denotes the matrix multiplication operation.\\
\indent Rather than computing independent spatial attention per modality, QKST explicitly transfers the attention map $\mathcal{A}_{t,l}^{opt}$ to the IR and SAR branches, guiding their structural alignment to OPT branch. The output feature $\hat{\mathbf{f}}_{t,l}^{m}$ of the subsequent attention block is thus given by:
\begin{equation}
\label{20}
\hat{\mathbf{f}}_{t,l}^{m} = \mathcal{A}_{t,l}^{opt} \otimes \mathbf{V}_{t,l}^{m}, \quad m \in \{opt, ir, sar\},
\end{equation}
where $\mathbf{V}_{t,l}^m$ is the value matrix in the UNet layer.\\
\indent These structure-aligned features $\hat{\mathbf{f}}_{t,l}^m$ are then propagated through the remaining UNet layers to produce the refined noise prediction $\hat{\epsilon}_\theta(\mathbf{z}_t^m, t, c)$. Following the diffusion iterative sampling rule, the latent state is updated to the next timestep $t-1$ to yield $\mathbf{z}_{T-1}^m$, as illustrated in Fig.~\ref{fig:QKST_inference}:
\begin{equation}
\label{21}
\mathbf{z}_{t-1}^m = \sqrt{\alpha_{t-1}}\left( \frac{\mathbf{z}_t^m - \sqrt{1-\alpha_t}\hat{\epsilon}_\theta}{\sqrt{\alpha_t}} \right) + \sqrt{1-\alpha_{t-1}}\hat{\epsilon}_\theta.
\end{equation}

By embedding the QKST mechanism into the iterative diffusion sampling pipeline, the model automatically regulates internal spatial structure under text guidance, avoiding the need for additional image supervision. This is achieved by the shared attention map $\mathcal{A}_{t,l}^{opt}$, which enforces a unified structural constraint across all modalities while still allowing each branch to adapt its modality attributes.
\section{EXPERIMENTS}
\subsection{EXPERIMENTS Setup}
\indent \textbf{Implementation details:} Our experiment is implemented using the PyTorch framework \cite{radford2021learning}, and model training and inference are evaluated on a single NVIDIA RTX 4090 GPU. We use SDv1.5 model\cite{rombach2022high} as the baseline backbone and fine-tune the UNet model on OPT remote sensing images for 10 epochs with a learning rate of 1e-5 to adapt the UNet to the remote sensing domain.
Subsequently, the UNet is frozen, and the CPD module is optimized using a two-stage training pipeline, as in Section \ref{dso}. The two stages are trained for 20 and 100 epochs, respectively, using a learning rate of 1e-4, batch size 8, and the Adam optimizer \cite{adam2014method} with momentum 0.9 and 0.99. During inference, denoising sampling is performed with the DDIM scheduler \cite{song2020denoising} using 50 steps, and the classifier-free guidance scale is set to 7.5.\\
\indent \textbf{Datasets:} In our experiments, we select the WHU-OPT-SAR large-scale multimodal paired remote sensing image dataset \cite{li2022mcanet} for training. This dataset includes covers six scene classes: $Farmland$, $City$, $Village$, $Water$, $Forest$, and $Road$. Each modality contains 7,000 images, split into 5,600 for training and 1,400 for testing, with all images cropped to 512 $\times$ 512 pixels. Additionally, we construct a small-target multimodal remote sensing image dataset (OSI) from several existing datasets. This OSI dataset includes eight different classes: $Beach$, $Desert$, $Lake$,  $Residential$, $Mountain$, $Farmland$, $River$, and $Ship$. The statistics of the OSI dataset are summarized in Table \ref{tab:dataset_stats}. Following an 8:2 train-test split, all images in this dataset are cropped to 256 $\times$ 256 pixels for training and testing. For quantitative evaluation, 4,200 images are generated from the WHU-OPT-SAR dataset and 3,360 images from the OSI dataset.\\
\indent \textbf{Evaluation metrics:} We select Fréchet Inception Distance (FID) \cite{heusel2017gans}, Inception Score (IS) \cite{salimans2016improved}, and CLIP score (CS) \cite{radford2021learning} metrics to evaluate model performance. The FID score is used to assess the distribution distance between generated images and real images in terms of modality attributes, with a smaller FID $\downarrow$ indicating that the quality of the generated images is closer to real images. The IS score is used to evaluate the diversity and clarity of the generated images, with a higher IS $\uparrow$ score indicating better quality of the generated images. Finally, we use the CLIP score to evaluate the alignment between text and images, with a higher CS $\uparrow$ indicating better semantic alignment from text to image. Through these evaluation metrics, we comprehensively validated the effectiveness and advancement of our proposed method.
\begin{table}[t] 
  \centering
  \caption{OSI Dataset Statistics}
  \label{tab:dataset_stats}
  \setlength{\tabcolsep}{1pt} 
  \resizebox{\linewidth}{!}{
  \begin{tabular}{lcccccccccc} 
    \toprule
    \multirow{2}{*}{\textbf{Dataset}} & \multirow{2}{*}{\textbf{Text}} & \multirow{2}{*}{\textbf{Image}} & \multicolumn{8}{c}{\textbf{Number of samples in different classes}} \\ 
    \cmidrule(lr){4-11} 
     & & & Beach. & Desert. & Lake. & Residential. & Mountain. & Farmland. & River. & Ship. \\ 
    \midrule
    $D_o^1$ & \ding{51} & \ding{51} & 700 & 700 & 700 & 700 & 700 & 700 & 700 & 700 \\
    $D_s^2$ & \ding{55} & \ding{51} & 812 & 828 & 44  & 59  & 808 & 796 & 565 & 1851 \\
    $D_I^3$ & \ding{55} & \ding{51} & 898 & 999 & 216 & 416 & 999 & 999 & 158 & 57 \\
    \bottomrule
    \multicolumn{11}{l}{\scriptsize $D^1_o$: NWPU-RESISC45 \cite{cheng2017remote}.} \\
    \multicolumn{11}{l}{\scriptsize $D^2_s$: MRSSC2.0 \cite{liu2022remote}, SARDet-100K \cite{li2024sardet}, BRIGHT \cite{su2024bright}, FUSAR-Ship\cite{hou2020fusar}.} \\
    \multicolumn{11}{l}{\scriptsize $D^3_I$: MRSSC2.0 \cite{liu2022remote}, DroneVehicle \cite{sun2022drone}, VEDAI \cite{razakarivony2014vehicle}.} \\
  \end{tabular}
  }
\end{table}
\begin{table*}[t]
\centering
\caption{Quantitative evaluation of remote sensing generation results. The best results are $\textbf{bolded}$ and the second-best results are
\underline{underlined}.}
\label{tab:sota_comparison_multimodal}

\definecolor{graybg}{gray}{0.90}
\setlength{\tabcolsep}{3pt}
\resizebox{0.96\textwidth}{!}{%
    \begin{tabular}{l | ccc ccc ccc | ccc ccc ccc}
    \toprule
    
    \multirow{3}{*}{\textbf{Methods}} & \multicolumn{9}{c|}{\textbf{WHU-OPT-SAR}} & \multicolumn{9}{c}{\textbf{OSI}} \\
    \cmidrule(lr){2-10} \cmidrule(l){11-19}
    
     & \multicolumn{3}{c}{OPT} & \multicolumn{3}{c}{IR} & \multicolumn{3}{c|}{SAR} 
     & \multicolumn{3}{c}{OPT} & \multicolumn{3}{c}{IR} & \multicolumn{3}{c}{SAR} \\
    \cmidrule(lr){2-4} \cmidrule(lr){5-7} \cmidrule(lr){8-10} 
    \cmidrule(lr){11-13} \cmidrule(lr){14-16} \cmidrule(l){17-19}
    
     & IS $\uparrow$ & FID $\downarrow$ & CS $\uparrow$ 
     & IS $\uparrow$ & FID $\downarrow$ & CS $\uparrow$ 
     & IS $\uparrow$ & FID $\downarrow$ & CS $\uparrow$
     & IS $\uparrow$ & FID $\downarrow$ & CS $\uparrow$ 
     & IS $\uparrow$ & FID $\downarrow$ & CS $\uparrow$ 
     & IS $\uparrow$ & FID $\downarrow$ & CS $\uparrow$ \\
    \midrule
    
    DF-GAN \cite{tao2022df} 
    & 3.056 & 43.03 & 0.2677 & 2.351 & 61.14 & 0.2676 & 2.598 & 51.26 & 0.2593 
    & 5.181 & 21.74 & 0.2541 & 1.741 & 21.31 & 0.2200 & 1.858 & 60.04 & 0.2321 \\ 
    
    OTD-GAN \cite{11126950}    
    & 2.367 & 39.71 & 0.2819 & 2.682 & 44.94 & \underline{0.2768} & 2.141 & 53.48 & 0.2673
    & 5.285 & 16.38 & 0.2552 & 1.633 & \underline{20.15} & 0.2258 & 1.845 & 38.97 & 0.2389 \\ 
    
    SDv1.5 \cite{rombach2022high}
    & 3.674 & 42.74 & 0.2785 & 3.734 & 39.66 & 0.2720 & \underline{3.303} & 58.65 & 0.2676 
    &5.359 & \underline{15.42}
 & \underline{0.2569}
 &  \underline{2.744} & 21.73 & 0.2363 & 1.933
 & 33.97 &  0.2461 \\
 
    CRS-Diff \cite{10663449}
    & 2.649 & 50.24 & 0.2658 & 4.086 & 42.74 & 0.2774 & 3.212 & 61.25 & \underline{0.2680}
    & \underline{5.622} & 20.94 & 0.2545 & 2.327 & 46.78 & 0.2408 & 1.948 & 56.51 & 0.2400 \\
    
    GeoSynth \cite{sastry2024geosynth}            
    & 3.176 & 51.32 & 0.2776 & 4.188 & 47.35 & 0.2701 & 2.725 & 58.75 & 0.2634
    & 5.083 & 16.72 & 0.2474 & 2.641 & 19.21 & 0.2440 & 2.146 & 59.10 & 0.2275 \\
    
    DiffusionSat \cite{khanna2024diffusionsat}
    & 2.743 & 41.55 & 0.2741 & 2.793 & 42.34 & 0.2735 & 3.222 & 57.59 & 0.2650 
    & 5.183 & 16.69 & 0.2472 & 2.743 & 21.73 & 0.2414 & \textbf{2.950} & 97.81 & 0.2350 \\
    
    Text2Earth \cite{10988859}
    & \underline{4.477} & \underline{28.43} & \underline{0.2844} & \underline{5.442} & \underline{42.12} & 0.2719 & 3.197 & \underline{49.90} & 0.2648
    & 5.278 & 16.54 & 0.2522 & 2.588 & 22.29 & \underline{0.2466} & 1.778 & \underline{33.09} & \underline{0.2474} \\
    \midrule
    \rowcolor{graybg}
    Ours   
     & \textbf{5.779}&\textbf{25.79} & \textbf{0.2886} & \textbf{5.904} &\textbf{27.31} & \textbf{0.2866} & \textbf{4.322}&\textbf{42.85} & \textbf{0.2855} 
    & \textbf{ 5.642
} & \textbf{12.84} & \textbf{0.2576} &  \textbf{2.935} & \textbf{17.64} & \textbf{0.2519} & \underline{2.713} & \textbf{18.74} & \textbf{0.2573} \\
    \bottomrule
    \end{tabular}%
}
\end{table*}
\begin{figure*}[t]
\includegraphics[width=0.96\textwidth]{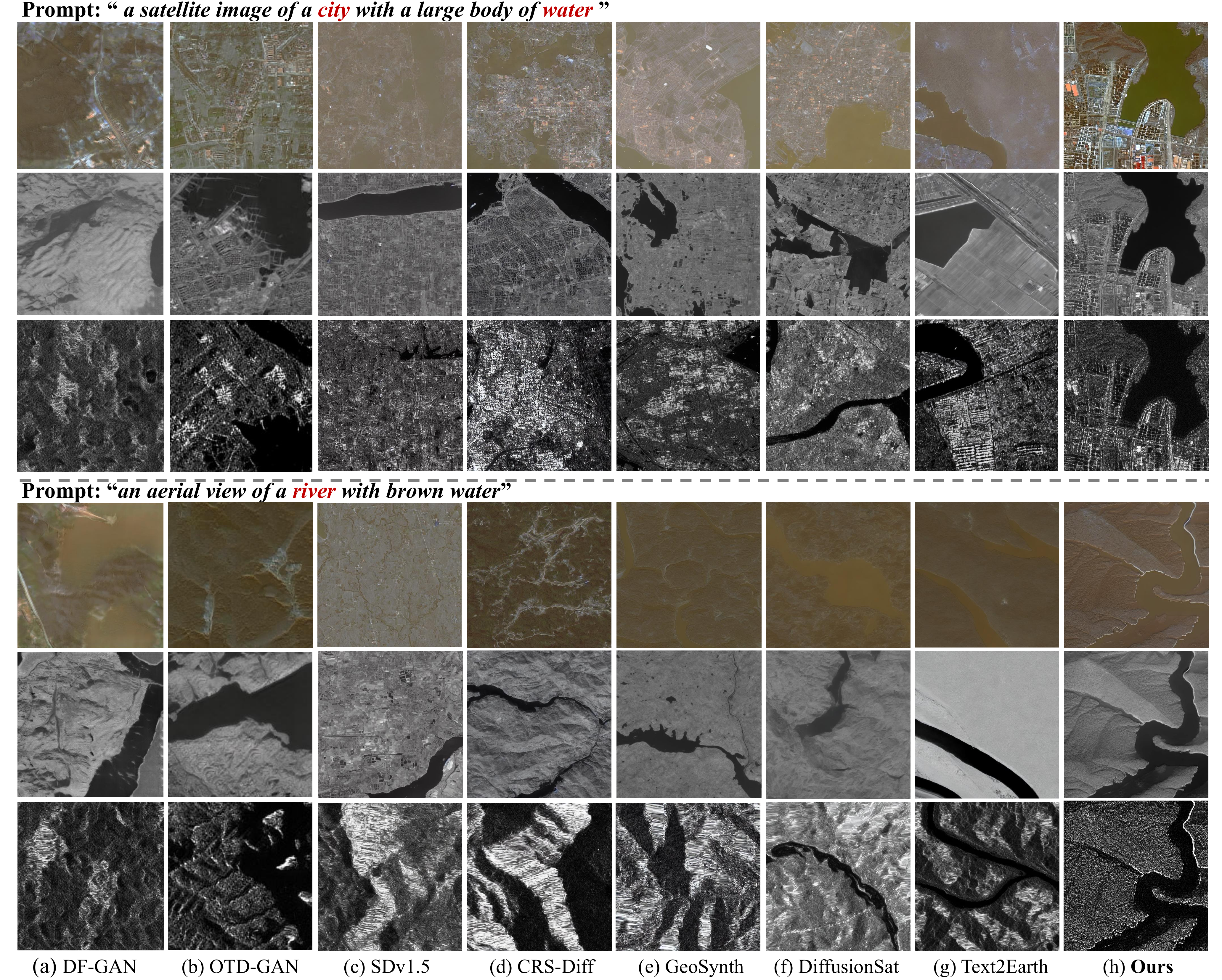}
		\centering \caption{Qualitative generation results on the WHU-OPT-SAR dataset. From top to bottom are the OPT, IR, and SAR modalities, respectively.}
\label{figcom}
        \vspace{-10pt}
	\end{figure*} 
\subsection{Comparison With State-of-The-Art Methods}
To evaluate the image quality generated by our proposed method, we compare it with previous remote sensing image generation methods from both qualitative and quantitative perspectives, including DF-GAN \cite{tao2022df}, OTD-GAN \cite{11126950}, SDv1.5 (Baseline) \cite{rombach2022high}, CRS-Diff\cite{10663449}, GeoSynth \cite{sastry2024geosynth}, DiffusionSat \cite{khanna2024diffusionsat}, and Text2Earth \cite{10988859}. We conduct comparative experiments on the WHU-OPT-SAR and OSI datasets.
As existing methods only support single-modal generation, we train each method individually across all three modalities on the basis of their official public implementations.
\begin{figure*}[t]
\includegraphics[width=0.96\textwidth]{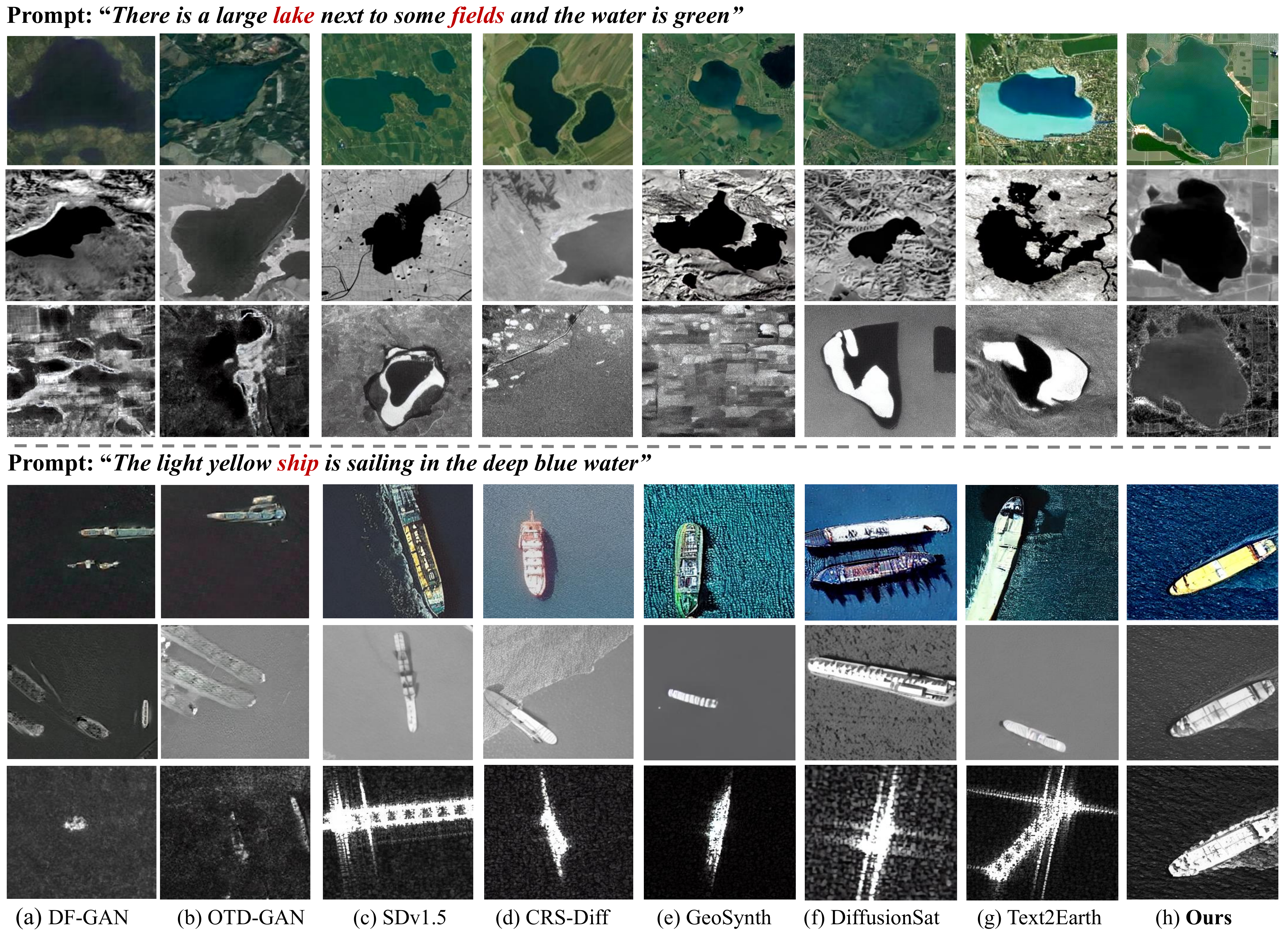}
		\centering \caption{Qualitative generation results on the OSI dataset. The generated images from top to bottom are the OPT, IR, and SAR modalities, respectively.}
\label{figcom1}
        \vspace{-5pt}
	\end{figure*} 
\begin{table*}[t]
\centering
\caption{Ablation study of different modules on the WHU-OPT-SAR and OSI datasets across three modalities, the \textbf{bold} indicates the best result.}
\label{TB2}
\definecolor{graybg}{gray}{0.90} 
\setlength{\tabcolsep}{3.5pt} 
\resizebox{\textwidth}{!}{%
    \begin{tabular}{l ccc ccc ccc ccc ccc ccc} 
    \toprule
    
    \multirow{3}{*}{\textbf{Method}} & \multicolumn{9}{c}{\textbf{WHU-OPT-SAR}} & \multicolumn{9}{c}{\textbf{OSI}} \\
    \cmidrule(lr){2-10} \cmidrule(l){11-19} 
    
     & \multicolumn{3}{c}{OPT} & \multicolumn{3}{c}{IR} & \multicolumn{3}{c}{SAR} 
     & \multicolumn{3}{c}{OPT} & \multicolumn{3}{c}{IR} & \multicolumn{3}{c}{SAR} \\
    \cmidrule(lr){2-4} \cmidrule(lr){5-7} \cmidrule(lr){8-10} 
    \cmidrule(lr){11-13} \cmidrule(lr){14-16} \cmidrule(l){17-19} 
    
     & IS $\uparrow$ & FID $\downarrow$ & CS $\uparrow$ 
     & IS $\uparrow$ & FID $\downarrow$ & CS $\uparrow$ 
     & IS $\uparrow$ & FID $\downarrow$ & CS $\uparrow$
     & IS $\uparrow$ & FID $\downarrow$ & CS $\uparrow$ 
     & IS $\uparrow$ & FID $\downarrow$ & CS $\uparrow$ 
     & IS $\uparrow$ & FID $\downarrow$ & CS $\uparrow$ \\
    \midrule
    
    Baseline          
    & \underline{3.674} & 42.74 & \underline{0.2785} 
    & 3.734 & 39.66 & 0.2720 
    & 3.303 & 58.65 & 0.2676 
    & \underline{5.359} & \underline{15.42} & \underline{0.2569} 
    & 2.744 & 21.73 & 0.2363 
    & 1.933 & 33.97 & 0.2461 \\
    
    Baseline+LoRA     
    & 3.666 & \underline{35.55} & 0.2722 
    & 4.775 & 44.90 & \underline{0.2826} 
    & 3.212 & 59.55 & 0.2726 
    & 5.355 & 15.50 & 0.2552 
    & 2.512 & 23.50 & 0.2512 
    & 2.546 & 22.29 & \underline{0.2506} \\
    
    Baseline+CPD      
    & 5.779 & 25.79 & 0.2886 
    & \textbf{6.048} & \underline{28.11} & 0.2823 
    & \textbf{5.335} & \textbf{42.81} & \underline{0.2734} 
    & 5.642 & 12.84 & 0.2576 
    & \textbf{2.950} & \underline{18.42} & \underline{0.2514} 
    & \underline{2.644} & \textbf{17.52} & 0.2504 \\
    
    \rowcolor{graybg}
    Baseline+CPD+QKST 
    & \textbf{5.779} & \textbf{25.79} & \textbf{0.2886} 
    & \underline{5.904} & \textbf{27.31} & \textbf{0.2866} 
    & \underline{4.322} & \underline{42.85} & \textbf{0.2855} 
    & \textbf{5.642} & \textbf{12.84} & \textbf{0.2576} 
    & \underline{2.935} & \textbf{17.64} & \textbf{0.2519} 
    & \textbf{2.713} & \underline{18.74} & \textbf{0.2573} \\
    
    \bottomrule
    \end{tabular}%
}
\end{table*}
\begin{figure}[!t]
\includegraphics[width=0.49\textwidth]{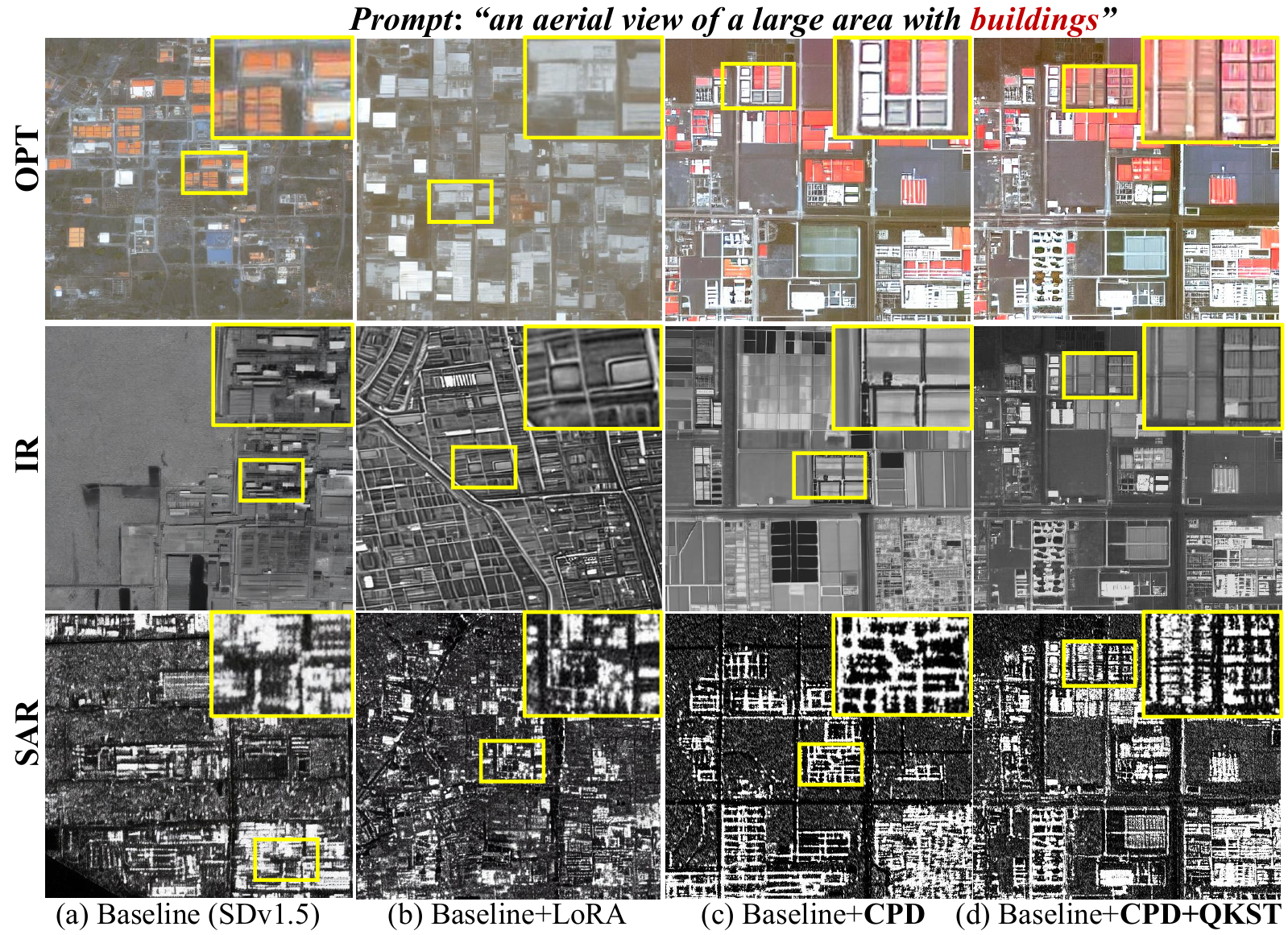}
		\centering 
		\caption{{Comparison generated results using LoRA \cite{hu2022lora} on the baseline model \cite{rombach2022high}, our proposed CPD module, and the QKST mechanism.}}
        \vspace{-10pt} 
		\label{fig8}
	\end{figure}
    \begin{figure}[!t]
\includegraphics[width=0.49\textwidth]{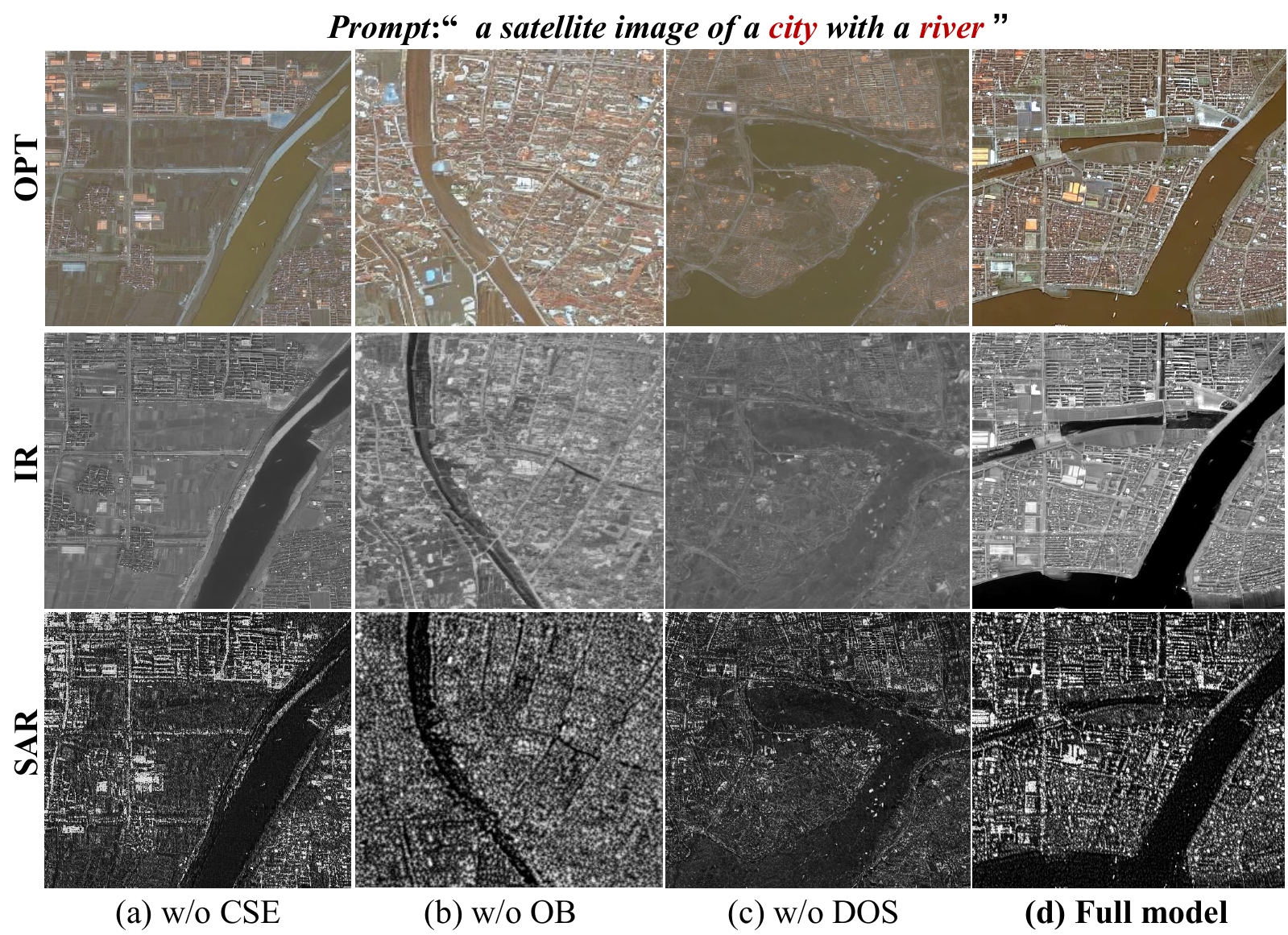}
		\centering 
		\caption{{Comparison results of different variants of CPD module.}}
        \vspace{-5pt} 
		\label{fig10}
	\end{figure}
    \begin{figure*}[t]
    \centering
    \begin{subfigure}[b]{0.48\linewidth}
        \centering
        \includegraphics[width=\linewidth]{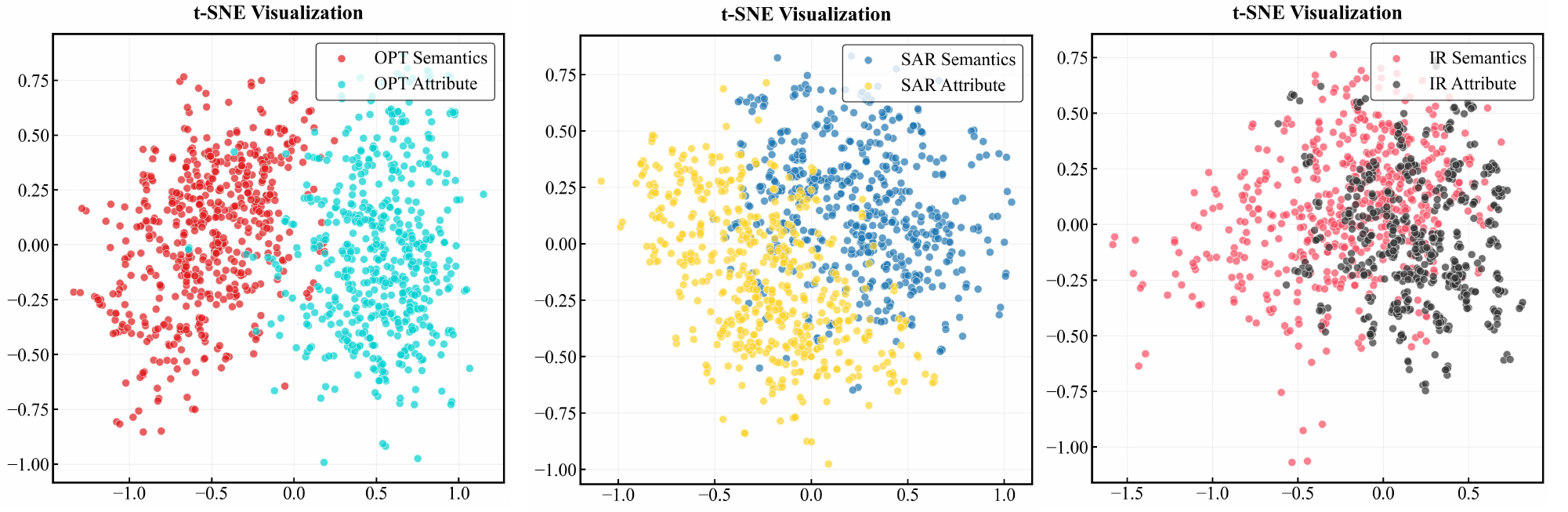}
        \caption{LoRA optimization paradigm }
        \label{fig:stability_analysis1}
    \end{subfigure}
    \hfill 
    \begin{subfigure}[b]{0.48\linewidth}
        \centering
        \includegraphics[width=\linewidth]{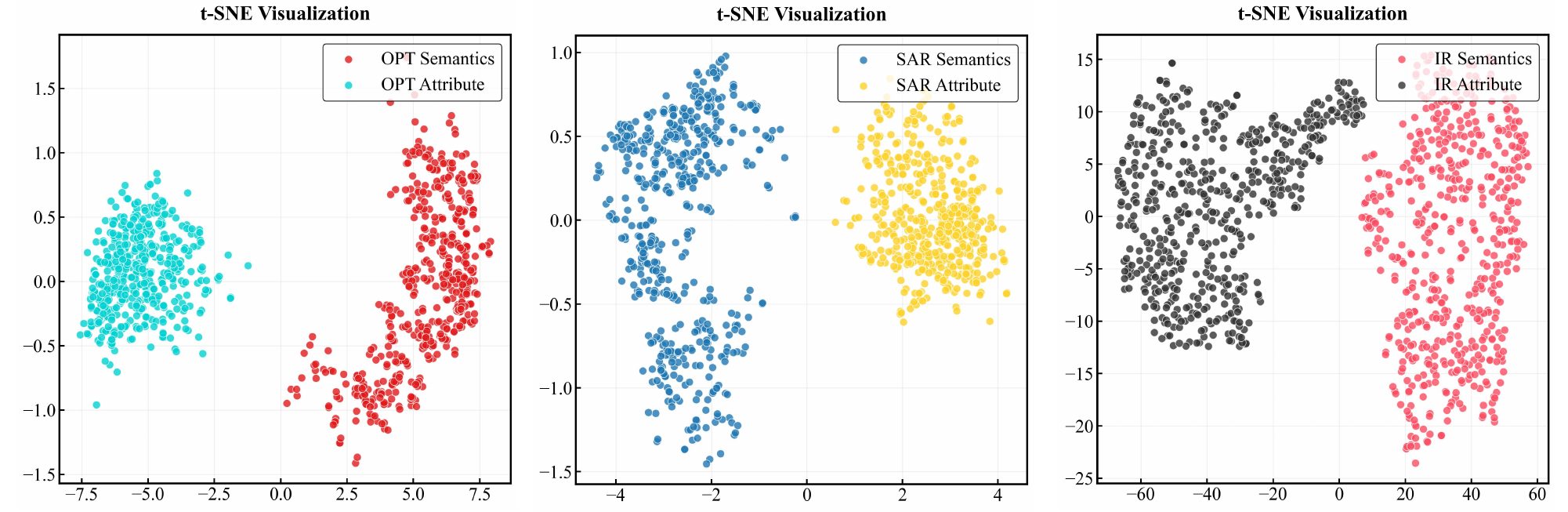}
        \caption{CPD optimization paradigm}
        \label{sde}
    \end{subfigure}
        \hfill
    \caption{t-SNE visualization of the feature disentanglement. (a) The conventional LoRA optimization paradigm \cite{hu2022lora} results in severe semantic-attribute entanglement. (b) Our proposed CPD paradigm achieves semantics-attribute disentanglement, thus improving the generation quality.}
    \label{fig:combined_motivation} 
\end{figure*}
\subsubsection{Qualitative Comparison} To validate the superiority of our proposed method in multimodal image generation, we conduct qualitative comparisons between our method and previous methods on both the WHU-OPT-SAR (as shown in Fig. \ref{figcom}) and OSI datasets (as shown in Fig. \ref{figcom1}). For IR and SAR modalities, previous methods produce images with compromised fine-grained details and texture fidelity. Specifically, generated IR images (columns a, b, f, g) fail to preserve subtle thermal patterns, while generated SAR images exhibit indistinct speckle textures and unrecognizable urban structures across all baselines. For IR and SAR modalities, previous methods produce images with compromised fine-grained details and texture fidelity. Specifically, IR images (columns a, b, f, g) fail to preserve rich detail patterns, while SAR images exhibit indistinct speckle textures and unrecognizable urban structures. Moreover, previous methods all cannot to produce spatially structurally aligned multi-modal images. Similar observations are obtained on the small-object OSI dataset (visualized in Fig.~\ref{figcom1}). For prompts involving small objects (e.g., ship) or structured scenes (e.g., large lake adjacent to fields), existing methods struggle to generate high-fidelity multimodal images with consistent semantics and aligned structures. In contrast, our method generates high-contrast OPT, detail-rich IR, and clear-textured SAR images from a single text prompt,
achieving favorable generation quality in both large-scale and object-scale remote sensing scenarios
while maintaining cross-modal structural alignment.
\subsubsection{Quantitative Evaluation} Table \ref{tab:sota_comparison_multimodal} presents the quantitative comparison of our method and previous remote sensing generation methods on the WHU-OPT-SAR and OSI datasets. Overall, our method consistently outperforms state-of-the-art approaches, attaining the best IS, FID, and CS across all modalities on the WHU-OPT-SAR dataset and leading performance in FID and CS on the OSI dataset. On the WHU-OPT-SAR dataset, the IS and FID improvement are attributed to our CPD module. It disentangles invariant semantics and modality attributes within the orthogonal core subspace, suppresses redundant noise, and enhances modality adaptation. Meanwhile, the remarkable CS improvements on IR and SAR modalities are attributed to our QKST mechanism. It explicitly aligns the structural characteristics of non-OPT modalities with the OPT anchor, strengthens cross-modal correspondence, and thus leads to substantial gains in the CS metric. On the OSI dataset, DiffusionSat \cite{khanna2024diffusionsat} achieves the marginally highest IS on the SAR modality, yet yields the worst FID. This outcome stems from a critical limitation where over-reliance on OPT priors generates semantically inconsistent SAR features that artificially inflate the IS while introducing severe distribution mismatch. Overall, these results confirm that our model excels at generating high-fidelity multimodal images with strong text-image alignment, especially for the challenging IR and SAR modalities.
\begin{figure}[!t]
\includegraphics[width=0.48\textwidth]{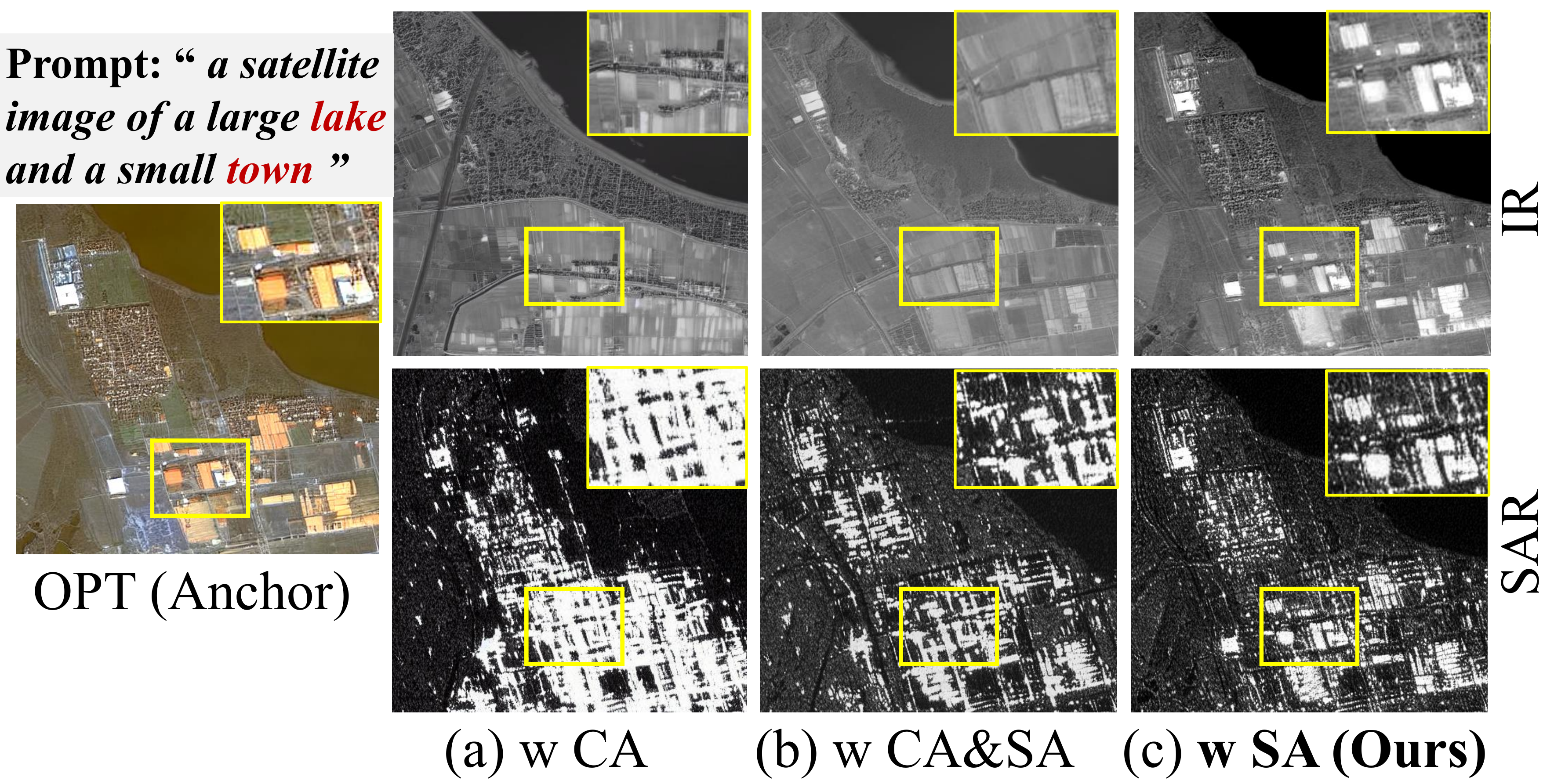}
		\centering 
		\caption{{Comparison results of different variants of QKST.}}
        \vspace{-5pt} 
		\label{fig11}
	\end{figure}
\subsection{Ablation Study}
\subsubsection{Effectiveness of the CPD module and QKST} In this work, we propose the CPD module, which maps invariant semantics from a single text prompt while adapting to distinct modality attributes. To validate its effectiveness, we individually integrate the LoRA adapter \cite{hu2022lora} and our CPD module into the baseline \cite{rombach2022high}, comparing the generated multimodal images and quantitative performance metrics. We first present a qualitative comparison of multimodal image generation results following the integration of distinct adapter modules into the baseline model. As shown in Fig.~\ref{fig8}(a)-(b), both the baseline and baseline+LoRA produce low-contrast, low-fidelity OPT images, detail-deficient IR images, and SAR images with blurry surface textures. In contrast, integrating our CPD module into the baseline enables simultaneous generation of high-fidelity, high-contrast OPT images, detail-rich IR images, and SAR images with clear textures from a single text prompt ($e.g.,$ the building highlighted by the yellow box in Fig.~\ref{fig8}(c)). Notably, standard LoRA adapters perform unconstrained low-rank perturbations and fail to explicitly disentangle semantics from modality attributes in multimodal generation. This is causing semantics-attributes entanglement and limiting the upper bound of generation quality. Conversely, our CPD module performs disentanglement within an orthogonal subspace and explicitly separates invariant semantics from modality attributes. It enables simultaneous adaptation to all three modalities under a single text prompt, yielding high generation fidelity and consistent semantics.\\
\indent In addition, we devise a QKST mechanism to ensure structural alignment between the generated multimodal images. To validate its effectiveness, we integrate QKST into the inference pipeline and evaluate the qualitative generation results. As illustrated in Fig.~\ref{fig8}(d), compared with the CPD-only results in Fig.~\ref{fig8}(c), the QKST mechanism effectively maintains consistent spatial structure across all generated multimodal images ($e.g.,$ the building highlighted by the yellow box in Fig.~\ref{fig8}(d)). This is because the QK attention maps in self-attention blocks explicitly encode the spatial structural relationships that dominate the diffusion trajectory (as shown in Fig. \ref{qkst}). By regulating these key spatial dependencies during inference, QKST aligns the denoising trajectories of the IR and SAR branches with the OPT anchor branch. Such diffusion trajectories preserve unified spatial structure across all modalities, ensuring cross-modal structural alignment as shown in Fig.~\ref{fig8}(d).\\
\begin{table}[t]
\centering
\caption{Exploration of different variants of CPD module on the WHU-OPT-SAR dataset. The \textbf{bold} indicates the best result.}
\label{tab:ablation_components}
\definecolor{graybg}{gray}{0.90}
\begin{tabular}{ll ccc}
\toprule
\textbf{Modality} & \textbf{Method} & \textbf{IS} $\uparrow$ & \textbf{FID} $\downarrow$ & \textbf{CS} $\uparrow$ \\
\midrule
\multirow{3}{*}{\textbf{OPT}} 
& w/o CSE & 4.442 & 27.42 & 0.2875 \\
& w/o OB  & 5.043 & 30.57 & 0.2839 \\
& w/o DOS & 4.530 & 40.05 & 0.2865 \\
\rowcolor{graybg}
& \textbf{Full model} & \textbf{5.779} & \textbf{25.79} & \textbf{0.2886} \\
\midrule
\multirow{3}{*}{\textbf{IR}}  
& w/o CSE & 4.810 & 34.97 & 0.2837 \\
& w/o OB  & 5.000 & 36.79 & 0.2845 \\
& w/o DOS & 4.081 & 34.38 & 0.2821 \\
\rowcolor{graybg}
&\textbf{Full model} & \textbf{5.904} & \textbf{27.31} & \textbf{0.2866} \\
\midrule
\multirow{3}{*}{\textbf{SAR}} 
& w/o CSE & 3.070 & 46.71 & 0.2648 \\
& w/o OB  & 3.946 & 57.89 & 0.2836 \\
& w/o DOS & 3.511 & 58.63 & 0.2787 \\
\rowcolor{graybg}
& \textbf{Full model} & \textbf{4.322} & \textbf{42.85} & \textbf{0.2855} \\
\bottomrule
\end{tabular}
\end{table}
\indent As shown in Table \ref{TB2}, we also conduct the ablation study on the WHU-OPT-SAR and OSI datasets to evaluate the performance of the above different modules when embedded into the baseline model. Upon integrating the LoRA adapter, the model yields competitive results across FID, IS, and CS metrics, showing a moderate improvement over the baseline. In stark contrast, integrating our CPD module into the baseline delivers substantial performance gains, achieving state-of-the-art IS/FID scores with an average relative improvement of $31.45\%$ and reduction of $27.27\%$ over the baseline (+LoRA) on both remote sensing datasets while preserving comparable or superior CS scores to the baseline. Furthermore, after embedding the QKST mechanism, the model maintains comparable performance across all evaluation metrics, demonstrating that it ensures cross-modal spatial structural alignment.
\begin{table}[t]
\centering
\caption{Exploration of QKST variants on the WHU-OPT-SAR dataset, where the OPT modality is adopted as the anchor to provide structural priors for IR and SAR modalities. The \textbf{bold} indicates the best result.}
\label{tab:ablation_step}
\definecolor{graybg}{gray}{0.90}
\begin{tabular}{ll ccc}
\toprule
\textbf{Modality} & \textbf{Method} & \textbf{IS} $\uparrow$ & \textbf{FID} $\downarrow$ & \textbf{CS} $\uparrow$ \\
\midrule
\multirow{3}{*}{\textbf{IR}} 
& w CA & 5.904 & 27.42 & 0.2857 \\
& w CA \& SA  & 5.885 &27.45 & 0.2839 \\
\rowcolor{graybg}
& \textbf{w SA (Ours)} & \textbf{5.904} & \textbf{27.31} & \textbf{0.2866} \\
\midrule
\multirow{3}{*}{\textbf{SAR}} 
& w CA & 4.087 & 47.46 & 0.2825 \\
& w CA \& SA  & \textbf{4.329} & 46.52 & 0.2814 \\
\rowcolor{graybg}
& \textbf{w SA (Ours)} & 4.322 & \textbf{42.85} & \textbf{0.2766} \\
\bottomrule
\end{tabular}
\end{table}
\subsubsection{Exploration of different CPD module's variants} We further conduct ablation exploration on different structural variants of the CPD module to verify the effectiveness and rationality of the current structural design. Specifically, we first investigate the necessity of core parameter space extraction (w/o CSE) by replacing it with orthogonal bases obtained from the full parameter space. As shown in Table~\ref{tab:ablation_components}, dropping this extraction step leads to consistent performance degradation across all modalities in terms of IS, FID, and CS. In particular, for the SAR modality, interference from redundant noise results in unclear content and degraded CS scores. This verifies that full parameters inevitably contain substantial redundancy accumulated during extensive training, while our core extraction mechanism effectively distills the most discriminative information and enhances the model’s generation capability. Qualitative results in Fig. \ref{fig10} corroborate these observations. As shown in Fig. \ref{fig10}(a), omitting the CSE operation yields darker multimodal generated images with reduced overall quality.  \\
\begin{table}[t]
\centering
\caption{Impact of different rank dimensions $r_1$ in the CPD.}
\label{tab:ablation_rank}
\definecolor{graybg}{gray}{0.90}
\begin{tabular}{ll ccc}
\toprule
\textbf{Modality} & \textbf{Method} & \textbf{IS} $\uparrow$ & \textbf{FID} $\downarrow$ & \textbf{CS} $\uparrow$ \\
\midrule
& $r_1=16$ & 5.729 & 31.11 & 0.2877 \\
\rowcolor{graybg}
& \textbf{$r_1=32$} & \textbf{5.779} & \textbf{25.79} & \textbf{0.2886} \\
\multirow{-3}{*}{\textbf{OPT}} & $r_1=64$ & 5.743 & 31.26 & 0.2867 \\ 
\midrule
& $r_1=16$ & \textbf{5.922} & 28.82 & 0.2852 \\
\rowcolor{graybg}
& \textbf{$r_1=32$} & 5.904 & \textbf{27.31} & \textbf{0.2866} \\
\multirow{-3}{*}{\textbf{IR}} & $r_1=64$ & 5.807 & 34.47 & 0.2855 \\ 
\midrule
& $r_1=16$ & \textbf{4.588} & 49.86 & 0.2853 \\
\rowcolor{graybg}
& \textbf{$r_1=32$} & 4.322 & \textbf{42.85} & \textbf{0.2855} \\
\multirow{-3}{*}{\textbf{SAR}} & $r_1=64$ & 3.938 & 42.98 & 0.2851 \\ 
\bottomrule
\end{tabular}
\end{table}
\indent We next remove the structural constraints of the orthogonal basis (w/o OB) and perform direct optimization in the core subspace using LoRA adapters with the disentangled optimization strategy.
As shown in Table~\ref{tab:ablation_components}, removing such orthogonal basis guidance leads to a substantial performance degradation across all multimodal generation metrics. This is because orthogonal bases maintain mathematical independence across distinct subspaces, which naturally regularizes the disentanglement of weight parameters corresponding to semantic and modality attributes, reduces interference during cross-modal adaptation, and thus improves the quality of generated images (as shown in Fig. \ref{fig10}(b), blurred urban texture details and distorted river boundary structures in generated images). \\
\indent We then ablate the disentangled optimization strategy of CPD (w/o DOS) by adopting the standard LoRA optimization paradigm \cite{hu2022lora} for model training.
As shown in Table \ref{tab:ablation_components}, all quantitative evaluation metrics of the model exhibit obvious degradation. To further interpret this phenomenon, we analyze the feature disentanglement behavior in Fig. \ref{fig:combined_motivation}(a). It can be observed that the traditional LoRA optimization paradigm fails to disentangle semantic attributes from modality attributes, making the model unable to learn clean and pure modality characteristics and thus leading to limited generation performance (as shown in Fig. \ref{fig10}(c)). In contrast, as illustrated in Fig. \ref{fig:combined_motivation}(b), our CPD paradigm achieves explicit and effective disentanglement between semantics and attributes, which fully preserves the inherent properties of each modality, thereby ensuring that the generated multimodal images have high contrast, rich details, and clear textures (as shown in Fig. \ref{fig10}(d)).
\begin{table*}[t]
\centering
\caption{Accuracy comparison of different methods in the downstream classification task across various training scales.}
\label{tab:comparison_wide}
\footnotesize 
\begin{tabular*}{0.95\textwidth}{@{\extracolsep{\fill}} l cccc cccc cccc @{}}
\toprule
\multirow{2}{*}{\textbf{Method}} & \multicolumn{4}{c}{\textbf{OPT}} & \multicolumn{4}{c}{\textbf{IR}} & \multicolumn{4}{c}{\textbf{SAR}} \\
\cmidrule(lr){2-5} \cmidrule(lr){6-9} \cmidrule(lr){10-13}
& 100\% & 200\% & 300\% & \textbf{Avg} & 100\% & 200\% & 300\% & \textbf{Avg} & 100\% & 200\% & 300\% & \textbf{Avg} \\
\midrule

\multicolumn{13}{c}{\textbf{Classification Backbone: VGG19}} \\
\midrule
DiffusionSat \cite{khanna2024diffusionsat} & 0.6116 & 0.6420 & 0.6795 & 0.6444 & 0.6060 & 0.6826 & 0.5048 & 0.5978 & 0.5144 & 0.5743 & 0.5803 & 0.5563 \\
Text2Earth \cite{10988859}   & 0.7071 & 0.7232 & 0.6589 & 0.6964 & 0.4679 & 0.6607 & 0.5075 & 0.5454 & 0.5384 & 0.6211 & 0.4388 & 0.5328 \\
OTD-GAN \cite{11126950}      & \textbf{0.8250} & 0.8268 & 0.7705 & 0.8074 & 0.4829 & 0.4528 & 0.4077 & 0.4478 & 0.4213 & 0.4555 & 0.4843 & 0.4537 \\
\rowcolor{gray!20} 
\textbf{Ours}       & \underline{0.8036} & \textbf{0.8580} & \textbf{0.8616} & \textbf{0.8411} & \textbf{0.6142} & \textbf{0.7005} & \textbf{0.6238} & \textbf{0.6128} & \textbf{0.5719} & \textbf{0.5995} & \textbf{0.5072} & \textbf{0.5595} \\

\midrule

\multicolumn{13}{c}{\textbf{Classification Backbone: ResNet50}} \\
\midrule
DiffusionSat \cite{khanna2024diffusionsat} & 0.6661 & 0.6071 & 0.7250 & 0.6661 & 0.5471 & 0.5498 & 0.6033 & 0.6001 & 0.4856 & 0.4664 & 0.5983 & 0.5168 \\
Text2Earth \cite{10988859}   & 0.7134 & 0.8214 & 0.7741 & 0.7696 & 0.5499 & 0.6019 & 0.4405 & 0.5308 & 0.4652 & 0.5036 & 0.4173 & 0.4620 \\
OTD-GAN \cite{11126950}      & 0.7705 & 0.8232 & 0.8018 & 0.7985 & 0.4049 & 0.4460 & 0.4596 & 0.4368 & 0.4257 & 0.4221 & 0.3705 & 0.4061 \\
\rowcolor{gray!20} 
\textbf{Ours}        & \textbf{0.8196} & \textbf{0.8670} & \textbf{0.8313} & \textbf{0.8393} & \textbf{0.6005} & \textbf{0.6156} & \textbf{0.6101} & \textbf{0.6087} & \textbf{0.6247} & \textbf{0.6031} & \textbf{0.6403} & \textbf{0.6227} \\

\bottomrule
\end{tabular*}
\end{table*}
\subsubsection{Exploration of different QKST's variants} 
We further conduct ablation studies on different transfer variants of the QKST mechanism to validate the efficacy and rationality of its design. Fig. \ref{fig11} presents the qualitative comparison of three QKST variants to validate the effectiveness of our current design: (a) transferring QK maps from the CA block in the UNet (w CA), (b) transferring QK maps from both CA and SA blocks (w CA\&SA), and (c) our design that transfers QK maps from only the SA block (w SA).\\
\indent Specifically, taking the OPT anchor image as the reference, the yellow boxes mark `town' with distinct spatial locations and structural layouts. For IR and SAR modalities, the CA-only transfer variant (Fig. \ref{fig11}(a)) fails to maintain precise spatial correspondence with the OPT anchor, leading to blurred building structures and misaligned layouts. The joint CA$\&$SA transfer variant (Fig. \ref{fig11}(b)) achieves partial improvement but still suffers from noticeable structural distortion and semantic misalignment. In contrast, our SA-only transfer design (Fig. \ref{fig11}(c)) faithfully reconstructs the exact spatial positions and detailed structural textures of the buildings in the yellow boxes, achieving cross-modal spatial structural consistency with the OPT anchor. This visual comparison confirms that exclusively transferring the SA block’s QK maps effectively captures cross-modal spatial semantic correspondences and enables precise spatial structure alignment, thus delivering optimal multimodal generation performance (as reported in Table~\ref{tab:ablation_step}).
\subsubsection{Impact of the different rank dimension} Finally, we perform an ablation study on the rank dimensions of parameter matrices $\mathbf{A}$ and $\mathbf{B}$ in the CPD module.
To achieve the optimal trade-off between generation performance and parameter efficiency, we explore three different settings with rank dimension $r_1 \in \{16, 32, 64\}$.
As shown in Table~\ref{tab:ablation_rank}, the model attains its best generation performance at $r_1 = 32$ while maintaining the low-rank efficiency of the parameter subspace.
Notably, $r_1 = 16$ provides insufficient representation capacity to capture complex radiometric textures, resulting in model underfitting.
In contrast, increasing $r_1$ to 64 brings no further performance improvement.
We attribute this to $r_1 = 64$ exceeding the intrinsic rank required for cross-modal learning, causing overfitting to useless patterns rather than learning generalizable feature distributions.
Such over-parameterization impairs the disentanglement performance of the CPD module and ultimately leads to degraded evaluation scores. Therefore, the rank dimension $r_1$ is set to 32. 
\subsection{Downstream Application in Image Classification}
To validate our proposed method as an efficient multimodal data generation engine for downstream remote sensing tasks, we conduct a quantitative evaluation under the train-on-synthetic, test-on-real protocol. This evaluation directly assesses the semantic fidelity and distribution consistency of our generated multimodal images by measuring their utility for downstream object classification. Specifically, we synthesize multi-scale datasets (100\%, 200\%, and 300\% of the original size) using text prompts from the validation set, ensuring generated samples maintain consistent semantics with real-world samples. We then train two standard classification backbones (VGG19 \cite{simonyan2014very} and ResNet50 \cite{he2016deep}) from scratch solely on these synthetic datasets. Classification accuracy on the unseen real test set serves as a direct metric to quantify the quality of generated samples and their generalization to real-world distributions.\\
\indent As presented in Table~\ref{tab:comparison_wide}, our method outperforms all competing methods across all modalities and synthesis scales, demonstrating its strong capability to generate high-quality multimodal images for downstream remote sensing classification tasks.
Notably, our method maintains the highest classification accuracy across all settings, showcasing more robust and stable adaptability for downstream tasks.
For instance, on SAR images with ResNet50, our method achieves 62.27\% accuracy, surpassing OTD-GAN by more than 20\%.
In contrast, existing methods exhibit noticeable performance fluctuations or even degradation across different modalities and synthesis scales. These results sufficiently validate that our proposed method can serve as an advanced and reliable data generation engine, which provides high-quality multimodal remote sensing images to effectively support downstream classification tasks.
\section{Conclusion}
This paper proposes the text-to-multimodal remote sensing image generation task, which synthesizes semantically and structurally consistent OPT, IR, and SAR images from a single text prompt. Distinct from existing single-modality methods, this task transcends current limitations by harnessing the complementary information inherent in multimodal images. To achieve this, we propose the CPD module, which leverages the functional dichotomy of LoRA adapters and performs dedicated disentangled optimization strategy within an orthogonal core subspace. By assigning parameter matrix A as a semantic anchor and parameter matrix B to model different modality attributes, our CPD module achieves explicit parameter-level disentanglement of semantics-attributes and thus enables high-fidelity, semantically consistent multimodal image generation from a single text prompt. We further devise the QKST mechanism during the inference process, which supports joint modeling of multimodal sampling trajectories for ensuring structural alignment across different modalities. Extensive experiments demonstrate the superior performance of our proposed method over previous state-of-the-art methods.
\bibliographystyle{IEEEtran}
\bibliography{bibf}
\vspace{-15mm}
\end{document}